\documentclass[letterpaper]{article} 
\usepackage{aaai2026}  
\usepackage{times}  
\usepackage{helvet}  
\usepackage{courier}  
\usepackage[hyphens]{url}  
\usepackage{graphicx} 
\usepackage{natbib}  
\usepackage{caption} 
\usepackage{algorithm}
\usepackage{algorithmic}
\usepackage{subfigure}
\usepackage{amsmath}
\usepackage{amssymb}
\usepackage{mathtools}
\usepackage{amsthm}
\usepackage{amsfonts,bm,amscd}
\usepackage{mathrsfs}
\newtheorem{theorem}{Theorem}

\newtheorem{assumption}[theorem]{Assumption}
\usepackage{multirow}
\usepackage{newfloat}
\usepackage{listings}
\DeclareCaptionStyle{ruled}{labelfont=normalfont,labelsep=colon,strut=off} 
\floatstyle{ruled}
\newfloat{listing}{tb}{lst}{}
\floatname{listing}{Listing}
\title{OT-ALD: Aligning Latent Distributions with Optimal Transport for Accelerated Image-to-Image Translation}
\author{
    Zhanpeng Wang\textsuperscript{\rm 1}, Shuting Cao\textsuperscript{\rm 1}, Yuhang Lu\textsuperscript{\rm 1}, Yuhan Li\textsuperscript{\rm 1}, Na Lei\textsuperscript{\rm 1}\thanks{Corresponding author. Email: nalei@dlut.edu.cn}, Zhongxuan Luo\textsuperscript{\rm 1}
}
\affiliations{
    \textsuperscript{\rm 1}Dalian University of Technology, Dalian, Liaoning, China
}

\usepackage{bibentry}

\begin{document}

\maketitle

\begin{abstract}
The Dual Diffusion Implicit Bridge (DDIB) is an emerging image-to-image (I2I) translation method that preserves cycle consistency while achieving strong flexibility. It links two independently trained diffusion models (DMs) in the source and target domains by first adding noise to a source image to obtain a latent code, then denoising it in the target domain to generate the translated image. However, this method faces two key challenges: (1) low translation efficiency, and (2) translation trajectory deviations caused by mismatched latent distributions. To address these issues, we propose a novel I2I translation framework, OT-ALD, grounded in optimal transport (OT) theory, which retains the strengths of DDIB-based approach. Specifically, we compute an OT map from the latent distribution of the source domain to that of the target domain, and use the mapped distribution as the starting point for the reverse diffusion process in the target domain. Our error analysis confirms that OT-ALD eliminates latent distribution mismatches. Moreover, OT-ALD effectively balances faster image translation with improved image quality. Experiments on four translation tasks across three high-resolution datasets show that OT-ALD improves sampling efficiency by 20.29\% and reduces the FID score by 2.6 on average compared to the top-performing baseline models.

\end{abstract}


\section{Introduction}
\label{sec:intro}
Image-to-image (I2I) translation is a fundamental task in computer vision that involves transforming an image from a source domain to a target domain while preserving its structural and semantic integrity. This technique has broad applications, including image restoration \cite{liang2021swinir,zamir2022restormer}, style transfer \cite{azadi2018multi,gatys2016image}, and image synthesis \cite{rombach2022high,odena2017conditional}, among others. The rapid advancement of deep generative models has led to significant progress in I2I translation, giving rise to numerous state-of-the-art methods. In particular, generative adversarial networks (GANs) \cite{zhu2017unpaired,isola2017image,yi2017dualgan,kim2017learning,fu2019geometry,park2020contrastive,benaim2017one} have played a pivotal role in early developments. More recently, diffusion models (DMs) \cite{sasaki2021unit,su2022dual,zhao2022egsde,saharia2022palette,wang2022pretraining,wang2022semantic,li2023bbdm,kwon2022diffusion,parmar2023zero,meng2021sdedit} have gained prominence and become a major focus in contemporary research due to their ability to produce high-quality, diverse outputs.

\begin{figure} [htbp]
    \centering
    \setlength{\abovecaptionskip}{0cm}
    \renewcommand{\figurename}{Fig}

    \subfigure[The translation effect of the proposed model on multiple high-resolution image tasks\label{fig:Toy_example:a}]{\includegraphics[width=1\columnwidth]{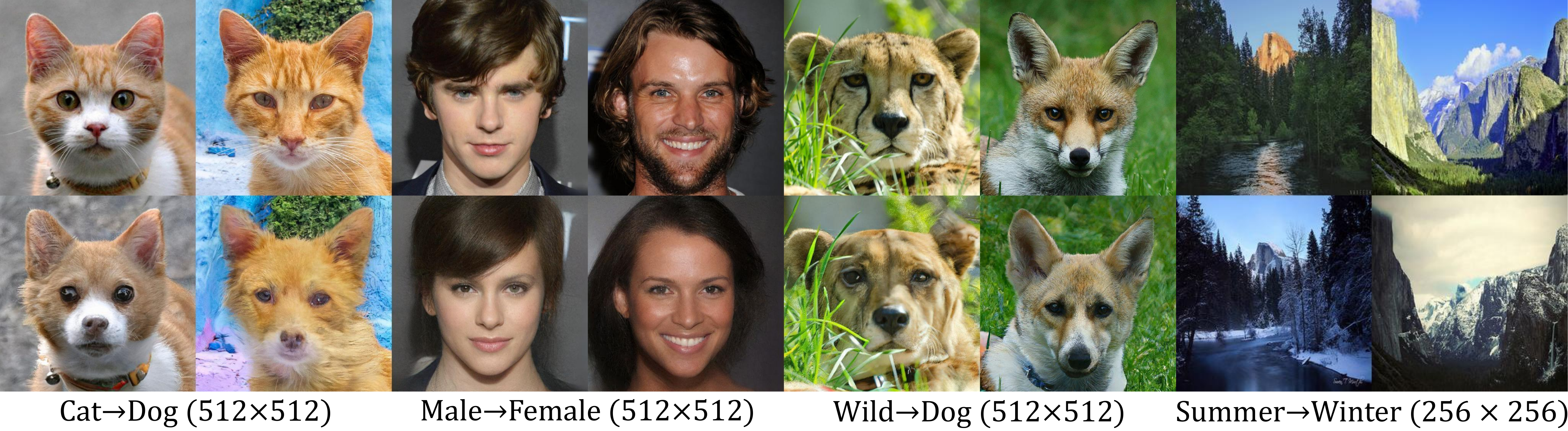}}\\
    \subfigure[Wasserstein difference in latent distributions\label{fig:Toy_example:b}]{\includegraphics[width=0.8\columnwidth]{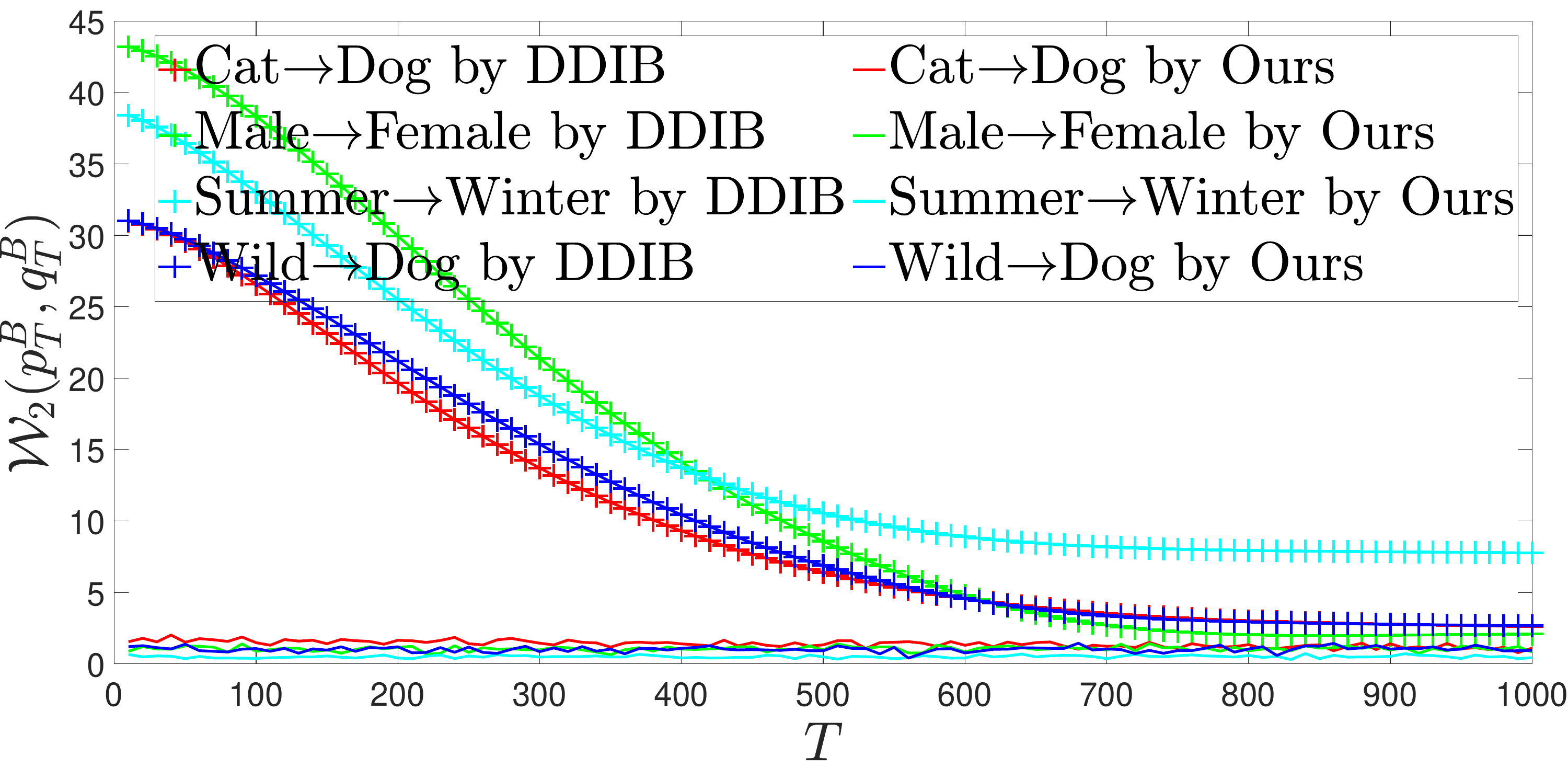}}
    \subfigure[Wasserstein difference in target distributions\label{fig:Toy_example:c}]{\includegraphics[width=0.8\columnwidth]{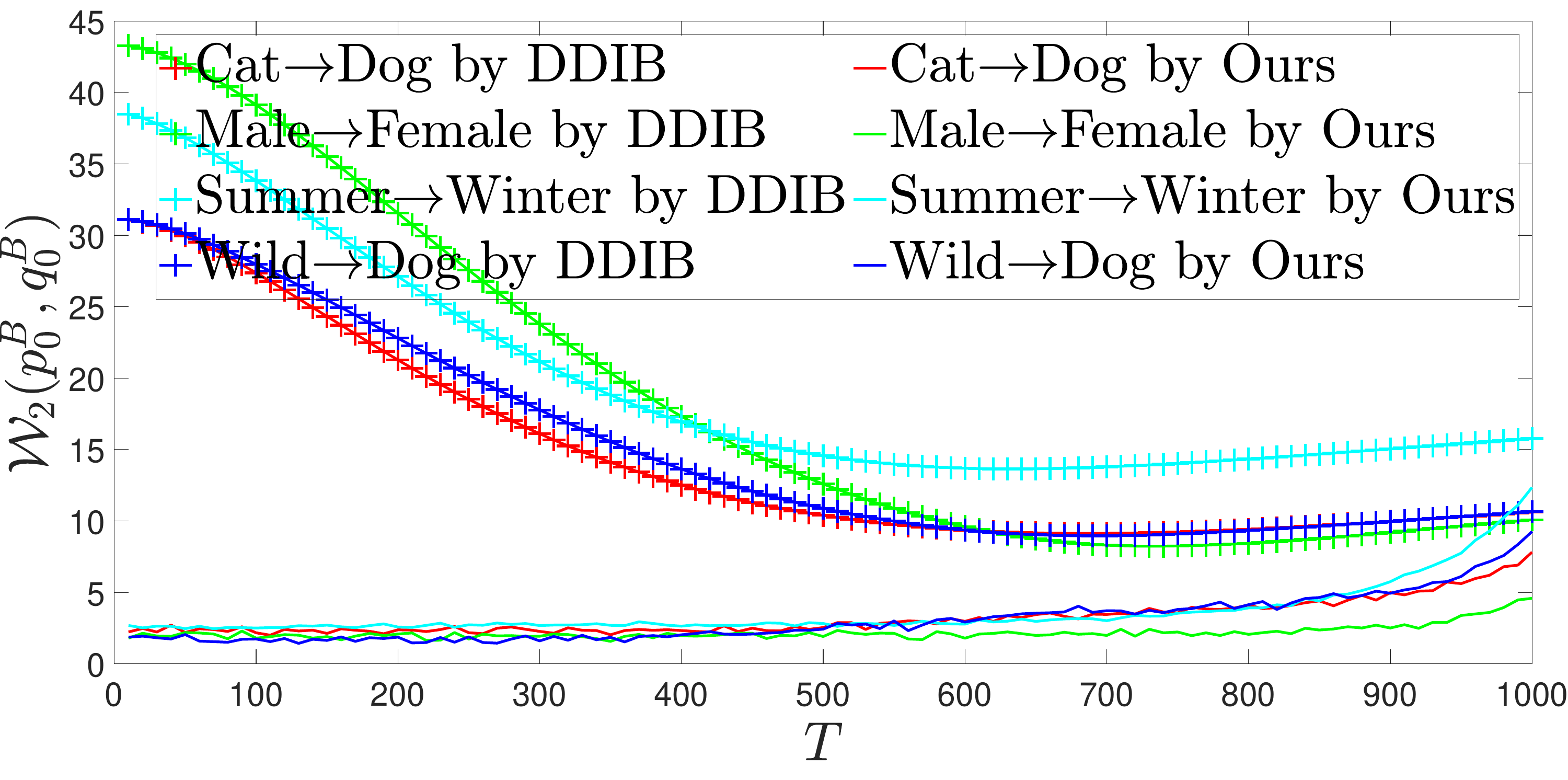}}
    \caption{(a) Top-row images are sources and bottom-row images are corresponding outputs generated by our model. (b) Contraction property of DMs. For initial distributions from different domains, DMs can exponentially shrink Wasserstein distance between their latent distributions through progressive noise injection. $T$ is the number of diffusion steps, and $q_{T}^{B}$ is initial distribution for the reverse process of $\mathrm{DM}^B$. In DDIB-based methods, $q_{T}^{B}=p_{T}^{A}$, while OT-ALD aligns latent distributions via OT map $M_{ot,T}^{A\to B}$ to ensure $q_{T}^{B}=M_{ot,T}^{A\to B}(p_{T}^{A})$. (c) Latent alignment impacts how closely the translated distribution approximates the ground truth (Theorem~\ref{theorem:Wasserstein_distance_upper_bound_DM}). DDIB requires longer diffusion to compensate for misalignment, reducing efficiency. OT-ALD is less sensitive to $T$, but empirically, too few steps harm image quality/diversity, and too many accumulate errors.}\label{Toy_example}
\end{figure}

The contraction property of DMs \cite{carrillo2006contractions,khrulkov2022understanding,franzese2023much}, as illustrated in Fig. \ref{fig:Toy_example:b}, significantly contributes to their rapid adoption in I2I translation. The core mechanism of DMs involves systematically corrupting a real image by progressively adding noise until it becomes pure Gaussian noise \cite{ho2020denoising}. Subsequently, the model learns to reconstruct the original image from noise through score matching \cite{song2020score}. This reconstruction is facilitated via a reverse process that incrementally removes noise, guiding the image from a random state back to the real data distribution. Mathematically, these forward and reverse processes are described by a pair of stochastic differential equations (SDEs) \cite{song2020score,anderson1982reverse}. Currently, diffusion-based I2I translation methods can be roughly categorized into two types: (1) Utilizing the source image to guide the generation trajectory of the target image \cite{zhao2022egsde,choi2021ilvr,meng2021sdedit,sun2023sddm,li2023injecting,lugmayr2022repaint,kim2022diffusionclip,tumanyan2023plug,seo2023midms,yang2023paint,zhang2023sine}, ensuring that the output evolves toward the desired target. However, this approach relies on joint training on both the source and target domains, necessitating simultaneous access to both datasets. This requirement limits data separation, undermines privacy protection, and reduces the overall flexibility. (2) Interconnecting two independently trained DMs in the source and target domains \cite{su2022dual,yu2023cross,zhang2023modeling,popov2023optimal,zhang2024decdm,bourou2024phendiff,hur2024expanding,yin2024scalable,mancusi2024latent}, a technique referred to as the Dual Diffusion Implicit Bridges (DDIB)-based method. Firstly, the source domain DM adds noise to the source image, transforming it into a latent code. The target domain DM then progressively denoises the latent code to generate the target image. This method enables I2I translation while ensuring privacy protection and high flexibility.

The DDIB-based method is not without its limitations. First, employing two DMs leads to increased inference time \cite{su2022dual}. Second, in practical applications, DMs must terminate after a finite number of time steps, meaning the termination distributions (latent code distributions) of the source and target domain DMs do not perfectly align with the standard Gaussian distribution. This misalignment introduces a theoretical gap, which has been shown to cause deviations in the image translation trajectory within the target domain. As a result, even with high-precision score matching training for both DMs, the translated images may fail to fully conform to the intended target distribution (see Theorem \ref{theorem:Wasserstein_distance_upper_bound_DM}). Furthermore, the cycle consistency of the DDIB-based method may also be compromised.

To address the abovementioned issues, we propose to eliminate this gap by aligning the latent code distribution using an optimal transport (OT) map (see Theorem \ref{corollary:Wasserstein_distance_upper_bound_OUR}). The main contributions of this paper are summarized as follows:

$ \bullet $ We demonstrate that existing DDIB-based I2I translation methods suffer from a theoretical gap caused by a mismatch between latent distributions. As a result, even with perfectly accurate score matching, the translation trajectory may still deviate from the intended target distribution.

$ \bullet $ Building on OT, we propose a novel I2I translation framework, OT-ALD, which theoretically aligns the latent distributions. Our theoretical analysis and experimental results show that OT-ALD inherits cycle consistency and strong flexibility of the DDIB-based approach.

$ \bullet $ OT-ALD can strike a balance between accelerating I2I translation and enhancing the quality of image generation. Experiments on four tasks across three high-resolution datasets show that compared with the best baseline model, OT-ALD achieves an average 20.29\% improvement in sampling efficiency and 2.6 reduction in FID.

\section{Preliminaries and related works}
\label{sec:Preliminaries and related works}
In this section, we first overview the background on image synthesis using DMs.  Next, we introduce the theoretical framework of OT.  Finally, we review the relevant literature related to this study.
\subsection{Image synthesis through the DMs}
The DMs include forward and reverse processes. Given the drift coefficient $ \boldsymbol{f}\left(\boldsymbol{x},t\right):\mathbb{R}^{n}\times\left[0,T\right]\to\mathbb{R}^{n} $ and the diffusion term $ g\left(t\right):\left[0,T\right]\to\mathbb{R}_{>0} $, the internal mechanism of the forward process can be elucidated by the following SDE \cite{song2020score}
\begin{equation}\label{eq:SDE}
    d\boldsymbol{x}=\boldsymbol{f}\left(\boldsymbol{x},t\right)dt+g\left(t\right)d\boldsymbol{W}_{t}, \boldsymbol{x}\left(0\right)\sim p_{0}\left(\boldsymbol{x}\right),
\end{equation}
where $ p_{0}\left(\boldsymbol{x}\right) $ is the image distribution and $ \boldsymbol{W}_{t} $ is the $ n $-dimensional standard Wiener process. The marginal distribution of the solution to \eqref{eq:SDE}, $ p_{t}\left(\boldsymbol{x}\right) $, satisfies the Fokker-Planck equation (FPE) \cite{risken1996fokker}
\begin{equation}\label{eq:Forward_Fokker_Planck}
    \frac{\partial p_{t}\left(\boldsymbol{x}\right)}{\partial t}+\nabla\cdot\left(\boldsymbol{f}\left(\boldsymbol{x},t\right)p_{t}\left(\boldsymbol{x}\right)\right)-\frac{g\left(t\right)^{2}}{2}\Delta p_{t}\left(\boldsymbol{x}\right)=0.
\end{equation}
According to \eqref{eq:Forward_Fokker_Planck}, there exists a deterministic process, referred to as the probability flow ordinary differential equation (PF-ODE) \cite{song2020score}, which shares the same marginal probability density $ p_{t} $ as \eqref{eq:SDE}
\begin{equation*}\label{eq:probability_flow_ODE}
	d\boldsymbol{x}=(\boldsymbol{f}\left(\boldsymbol{x},t\right)-\frac{g\left(t\right)^{2}}{2}\nabla\log p_{t}\left(\boldsymbol{x}\right))dt,\boldsymbol{x}\left(0\right)\sim p_{0}\left(\boldsymbol{x}\right).
\end{equation*}
Similarly, the reverse process can be expressed as reverse-time SDE \cite{anderson1982reverse} with $q_{T}$ as the initial condition
\begin{equation}\label{eq:reverse_time_SDEs}
    d\boldsymbol{x}=(\boldsymbol{f}\left(\boldsymbol{x},t\right)-g\left(t\right)^{2}\nabla\log p_{t}\left(\boldsymbol{x}\right))dt+g\left(t\right)d\bar{\boldsymbol{W}}_{t},
\end{equation}
where $\boldsymbol{x}\left(T\right)\sim q_{T}\left(\boldsymbol{x}\right)$ and $ \bar{\boldsymbol{W}}_{t} $ is the $ n $-dimensional standard Wiener process with backward time.
In practice, $ \nabla\log p_{t}\left(\boldsymbol{x}\right) $ is usually approximated by a score network $ \boldsymbol{S}_{\boldsymbol{\theta}}\left(\boldsymbol{x},t\right):\mathbb{R}^{n}\times\left[0,T\right]\to\mathbb{R}^{n} $ with the weighted mean square error (MSE) loss function \cite{song2021maximum}
{\small
\begin{equation*}\label{eq:weighted_mean_square_error}
    \mathcal{J}_{SM}\left(\boldsymbol{\theta},\phi,T\right):=\frac{1}{2}\int_{0}^{T}\phi\left(t\right)\mathbb{E}_{p_{t}}[\left \| \boldsymbol{S}_{\boldsymbol{\theta}}\left(\boldsymbol{x},t\right)-\nabla\log p_{t}\left(\boldsymbol{x}\right) \right \|_{2}^{2}]dt,
\end{equation*}}
here $ \phi\left(t\right):\left[0,T\right]\to\mathbb{R}_{>0} $ is a positive weighting function. By replacing $ \nabla\log p_{t}\left(\boldsymbol{x}\right) $ with $ \boldsymbol{S}_{\boldsymbol{\theta}}\left(\boldsymbol{x},t\right) $ in the reverse process, we get an approximate reverse-time SDE and FPE. Ultimately, the approximate image distribution $q_{0}$ is obtained from $q_{T}$. The solution of DMs can be formalized as
\begin{equation}\label{eq:solution_DMs}
    \begin{split}
        &Solver_{\eta}(\boldsymbol{x}_{t_{0}},\boldsymbol{S}_{\boldsymbol{\theta}},t_{0},t_{1})\\
        =&\boldsymbol{x}_{t_{0}}+\int_{t_{0}}^{t_{1}}\boldsymbol{v}_{\boldsymbol{\theta}}\left(\boldsymbol{x},t\right)dt+\eta\int_{t_{0}}^{t_{1}}g(t)d\boldsymbol{W}_{t},
    \end{split}
\end{equation}
where $\boldsymbol{v}_{\boldsymbol{\theta}}\left(\boldsymbol{x},t\right)=\boldsymbol{f}\left(\boldsymbol{x},t\right)-\frac{1}{2}(1-\eta^{2})g\left(t\right)^{2}\boldsymbol{S}_{\boldsymbol{\theta}}\left(\boldsymbol{x},t\right)$ and $\eta\in[0,1]$ denotes noise scaling factor.
\subsection{Optimal transport theory}\label{sec:OT}
We begin by introducing the classical geometric variational approach to semi-discrete OT. As the number of target samples increases, the discrete solution progressively converges to its continuous counterpart. Suppose the source measure $\mu$ is defined on a convex domain $\Omega \subset \mathbb{R}^{n}$, and the target domain is a discrete set $\mathcal{Y} = \{\boldsymbol{y}_i\}_{i \in \mathcal{I}}\subset\mathbb{R}^{n}$. The target measure is a Dirac measure $\nu = \sum_{i \in \mathcal{I}} \nu_i \delta(\boldsymbol{y} - \boldsymbol{y}_i)$, with the total mass matched as $\mu(\Omega) = \sum_{i \in \mathcal{I}} \nu_i$. Under a transport map $M: \Omega \rightarrow \mathcal{Y}$, the domain $\Omega$ is partitioned into cells $W_i$ such that each point $\boldsymbol{x} \in W_i$ is mapped to $\boldsymbol{y}_i$, i.e., $M(\boldsymbol{x}) = \boldsymbol{y}_i$. The map $M$ is measure-preserving (denoted $M_{\#}\mu = \nu$) if each cell satisfies $\mu(W_i) = \nu_i$. Let $c: \Omega \times \mathcal{Y} \rightarrow \mathbb{R}$ be the cost function, where $c(\boldsymbol{x}, \boldsymbol{y})$ denotes the cost of transporting unit mass from $\boldsymbol{x}$ to $\boldsymbol{y}$. The total transport cost of the map $M$ is given by
\begin{equation}
\int_{\Omega} c(\boldsymbol{x}, M(\boldsymbol{x})) d \mu(\boldsymbol{x})=\sum_{i\in\mathcal{I}} \int_{W_{i}} c\left(\boldsymbol{x}, \boldsymbol{y}_{i}\right) d \mu(\boldsymbol{x}).
\label{eq:cost}
\end{equation}
The OT map $M_{ot}$ is a measure-preserving map that minimizes the total cost in \eqref{eq:cost},
\begin{equation}
M_{ot}:=\arg \min _{M_{\#} \mu=\nu} \int_{\Omega} c(\boldsymbol{x}, M(\boldsymbol{x})) d \mu(\boldsymbol{x}).\label{eq:SDOT}
\end{equation}
Specifically, when the cost function is \( c(\boldsymbol{x}, \boldsymbol{y}) = \frac{1}{2}\|\boldsymbol{x} - \boldsymbol{y}\|_2^2 \), there exists a convex function \( u: \Omega \to \mathbb{R} \) such that its gradient \( \nabla u \) uniquely solves \eqref{eq:SDOT} \cite{brenier1991polar}, i.e., \( M_{ot} = \nabla u \). This shows that the OT map is the gradient of Brenier's potential. As noted in \cite{lei2020geometric,an2020ae}, \( u \) can be represented as the upper envelope of hyperplanes
$\pi_{\boldsymbol{h},i}(\boldsymbol{x}) = \langle \boldsymbol{x}, \boldsymbol{y}_i \rangle + h_i$,
and is uniquely parameterized (up to an additive constant) by a height vector \( \boldsymbol{h} = (h_1, h_2, \dots, h_{|\mathcal{I}|})^T \). The resulting potential, denoted \( u_{\boldsymbol{h}} \), is given by:
\begin{equation}
    u_{\boldsymbol{h}}(\boldsymbol{x}) = \max_{i\in\mathcal{I}}\{\pi_{\boldsymbol{h},i}(\boldsymbol{x})\}, u_{\boldsymbol{h}}: \Omega \rightarrow \mathbb{R}^n.\label{eq:uh}
\end{equation}
Given the target measure \( \nu \), there exists a Brenier potential \( u_{\boldsymbol{h}} \) (as in \eqref{eq:uh}) such that the projected volume under each support plane matches the prescribed mass \( \nu_i \). To obtain \( u_{\boldsymbol{h}} \), it suffices to optimize the height vector \( \boldsymbol{h} \) by minimizing the following convex energy function:
\begin{equation}
    E(\boldsymbol{h})=\int_{\boldsymbol{0}}^{\boldsymbol{h}} \sum_{i\in\mathcal{I}} w_{i}(\gamma) d \gamma_{i}-\sum_{i\in\mathcal{I}} h_{i} \nu_{i},
\label{eq:EH}
\end{equation}
where $\omega_{i}(\gamma)$ is the $\mu$-volume of $W_{i}(\gamma)$. For more theoretical details, we refer to Reference \cite{gu2013variational}.
\subsection{Related works}
This section presents a curated overview of relevant literature, with additional references available in \textit{Appendix A} and in the review articles \cite{hoyez2022unsupervised,huang2024diffusion}.


\textbf{Conditional guided and controlled translation.} To enhance control over image generation, researchers have proposed multimodal conditional strategies using text, exemplars, or semantic labels. Kim et al. \cite{kim2022diffusionclip} applied DMs with CLIP loss for text-guided editing, while Meng et al. \cite{meng2021sdedit} synthesized realistic images by adding noise to images and denoising via reverse-time SDEs. Tumanyan et al. \cite{tumanyan2023plug} enabled fine-grained control by manipulating internal features of pre-trained models. Seo et al. \cite{seo2023midms} improved exemplar-guided accuracy via alternating cross-domain matching and latent diffusion. Cheng et al. \cite{cheng2023general} addressed single-example translation through content-concept inversion and fusion. Classifier-free guidance was adopted in \cite{yang2023paint,zhang2023sine} for controlled editing. Kwon et al. \cite{kwon2022diffusion} performed text- and image-guided style transfer by aligning attention keys and class tokens, while Shi et al. \cite{shi2024dragdiffusion} introduced latent optimization with reference latent control for point-interactive editing.

\textbf{Efficient and lightweight translation.} To mitigate the high computational cost of DMs, various optimization strategies have been explored. Song et al. \cite{song2020denoising} introduced non-Markovian sampling to reduce generation steps, while Luo et al. \cite{luo2023latent} achieved single-step inference via consistency distillation. Xia et al. \cite{xia2024diffi2i} proposed compact prior networks and dynamic transformers, and Jiang et al. \cite{jiang2024fast,xia2024diffusion} optimized time-step usage to accelerate translation. Parmar et al. \cite{parmar2024one} combined single-step DMs with adversarial learning for fast, high-quality synthesis. Lee et al. \cite{lee2024ebdm} employed Brownian bridge strategies for better stability and efficiency. Additionally, Lee et al. \cite{lee2024diffusion} improved text-driven I2I translation by interpolating prompts to refine noise prediction in pre-trained DMs.

\section{Methodology of OT-ALD}
We use superscripts $^{A}$ and $^{B}$ to indicate their association with the source and target domains, respectively. Fig.\ref{framework} illustrates the essential differences between OT-ALD and DDIB-based methods \cite{su2022dual,yu2023cross,zhang2023modeling,popov2023optimal,zhang2024decdm,bourou2024phendiff,hur2024expanding,yin2024scalable,mancusi2024latent}. Given an initial image distribution $p_{0}^{A}$ in the source domain, DDIB-based methods advocate using the forward process of $\mathrm{DM}^{A}$ to corrode $p_{0}^{A}$ into a latent code distribution $p_{T}^{A}$. The reverse process of $\mathrm{DM}^{B}$ then reconstructs the target image distribution $p_{0}^{B}$ from $p_{T}^{A}$. According to the contraction property of SDEs, as illustrated in Fig.\ref{fig:Toy_example:b}, the Wasserstein distance $\mathcal{W}_{2}(p_{T}^{A}, p_{T}^{B})$ converges exponentially to zero as $T \to +\infty$ \cite{khrulkov2022understanding,franzese2023much}. However, in practical scenarios with finite $T$, we have $\mathcal{W}_{2}(p_{T}^{A}, p_{T}^{B}) \ne 0$, resulting in a mismatch between latent code distributions. Here we present Theorem \ref{theorem:Wasserstein_distance_upper_bound_DM}, which demonstrates that this mismatch leads to deviations in the translation trajectory within the target domain, ultimately preventing the attainment of the desired target distribution. The proof can be found in \textit{Appendix D.1}.

\begin{figure} [htbp]
    \centering
    \renewcommand{\figurename}{Fig}
    \includegraphics[width=1\columnwidth]{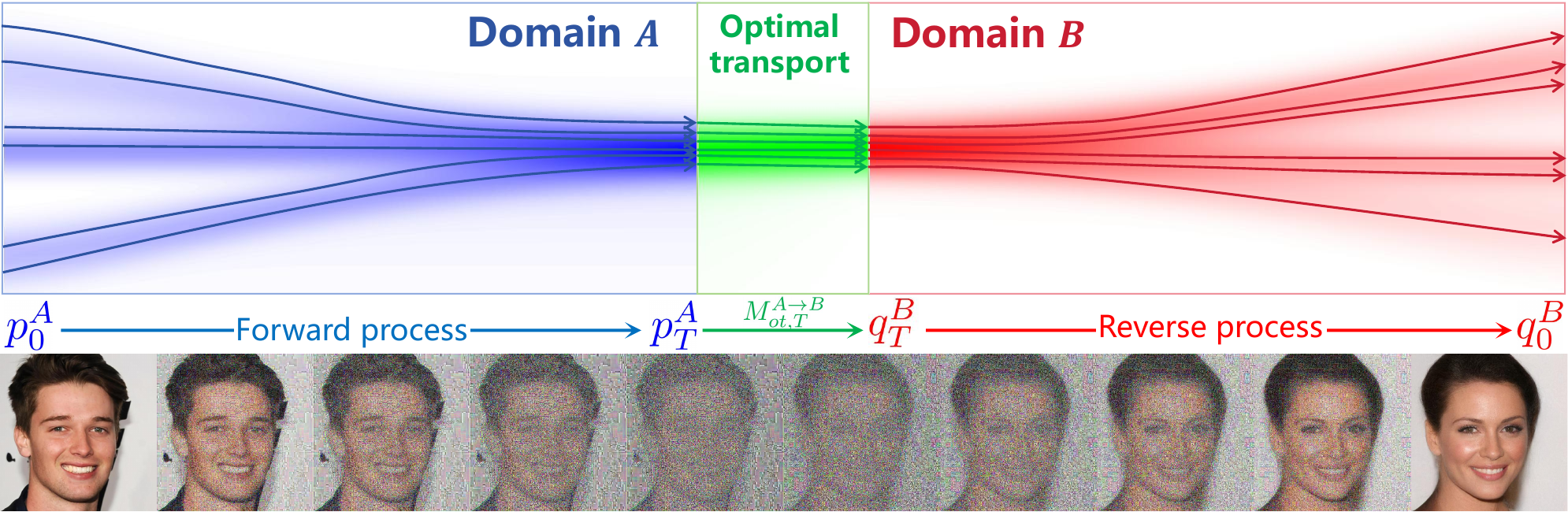}%
    \caption{The framework of OT-ALD. During training, two DMs are independently trained in domains $A$ and $B$, followed by computation of the OT map $M_{ot,T}^{A \to B}$ from $p_T^A$ to $p_T^B$. In translation, the source distribution $p_0^A$ is diffused to $p_T^A$, which is then mapped via $M_{ot,T}^{A \to B}$ to serve as the initial distribution for the reverse process in domain $B$, yielding the translated distribution $q_0^B$. In contrast, DDIB-based methods skip OT alignment and directly use $p_T^A$, leading to latent distribution mismatch. As shown in Theorem~\ref{theorem:Wasserstein_distance_upper_bound_DM}, this mismatch introduces a theoretical gap that affects translation accuracy.}\label{framework}
\end{figure}
\begin{figure} [htbp]
    \centering
    \renewcommand{\figurename}{Fig}
    \subfigure[$T=500,\eta:0\to1$.\label{fig:eta_T:a}]{\includegraphics[width=0.493\columnwidth]{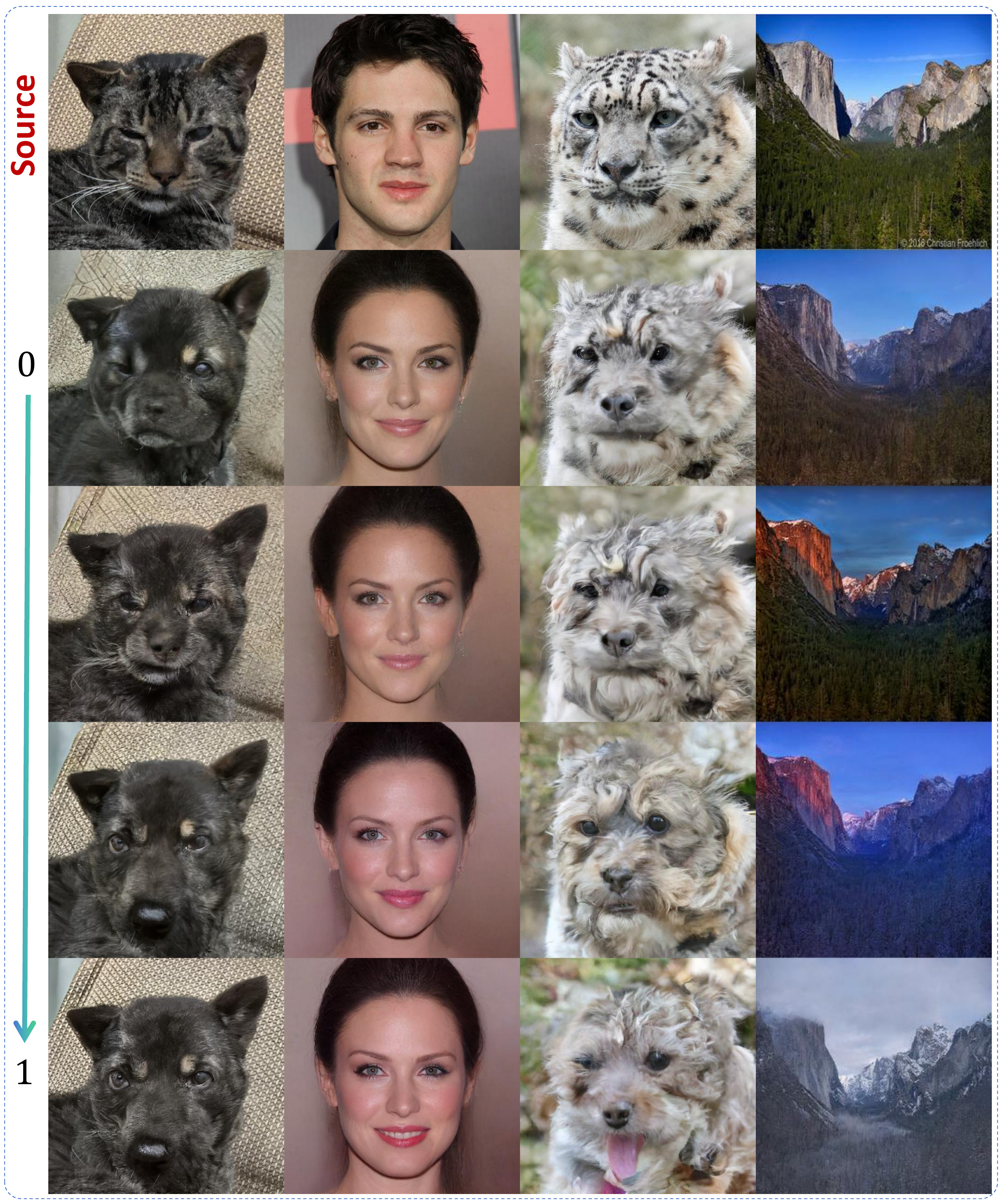}}
    \subfigure[$\eta=0.4$, $T/1000:0\to1$.\label{fig:eta_T:b}]{\includegraphics[width=0.493\columnwidth]{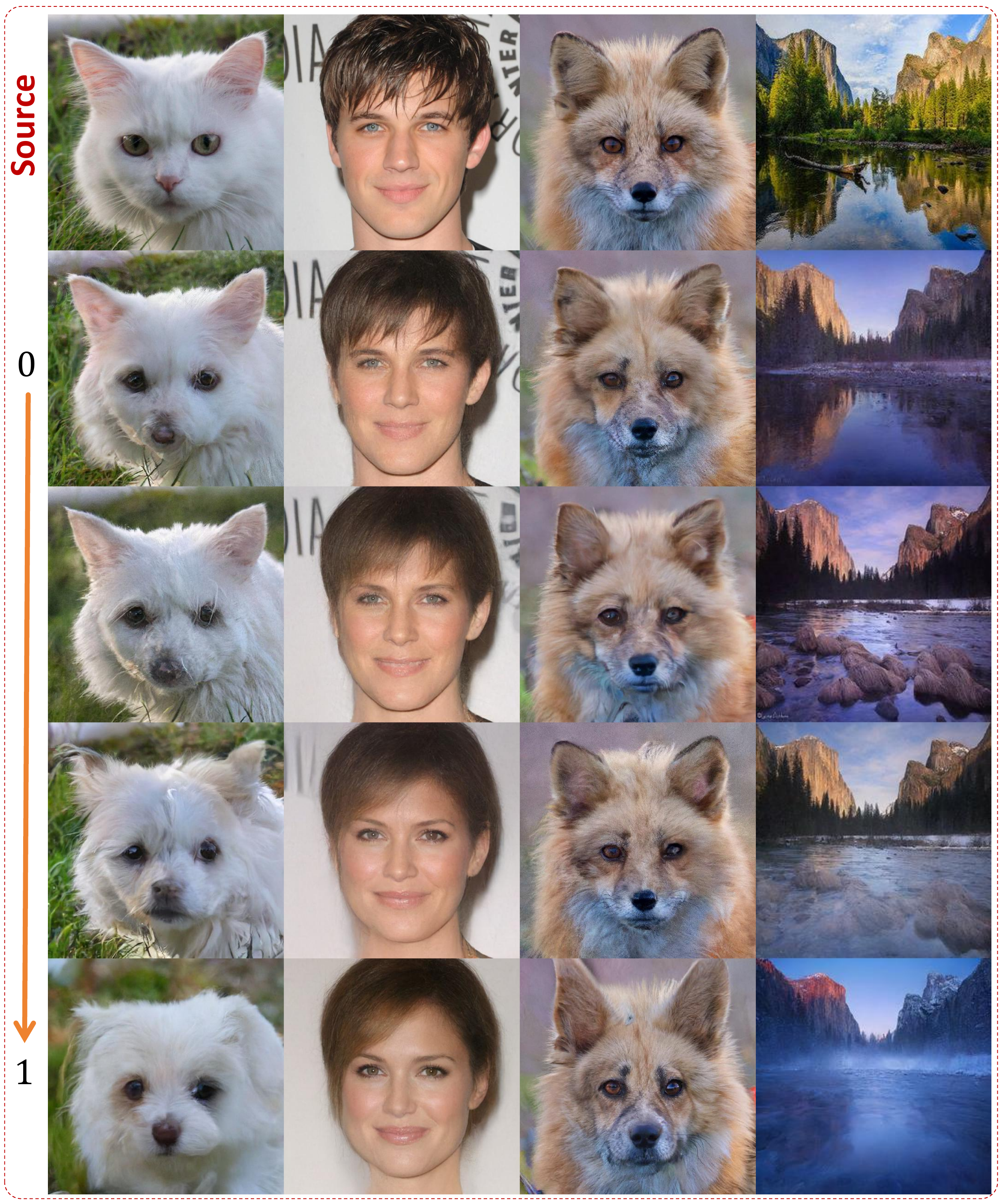}}
    \caption{The smaller the adopted $\eta$ and $T$, the higher the fidelity of OT-ALD to the source image is maintained;  Conversely , larger values of $\eta$ and $T$ indicate that OT-ALD can achieve greater diversity in I2I translation tasks.}\label{eta_T}
\end{figure}

\begin{theorem}\label{theorem:Wasserstein_distance_upper_bound_DM}
    If $ q_{T}^{B}=p_{T}^{A} $, we denote $ q_{0}^{B} $ as the distribution generated by the DM$^{B}$, then $ \mathcal{W}_{2}(p_{0}^{B},q_{0}^{B}) $ can be estimated as follows
    \begin{equation*}
        \bar{I}^{B}\left(T\right)\mathcal{W}_{2}(p_{T}^{A},p_{T}^{B})\leq\mathcal{W}_{2}(p_{0}^{B},q_{0}^{B})\leq I^{B}\left(T\right)\mathcal{W}_{2}(p_{T}^{A},p_{T}^{B}),
    \end{equation*}
    where $ \!I^{B}\left(T\right)=\exp\left(\int_{0}^{T}(L_{f}^{B}\left(t\right)+\frac{g^{B}\left(t\right)^{2}}{2}L_{\boldsymbol{S}_{\boldsymbol{\theta}}}^{B}\left(t\right))dt\right)\! $ and $ \!\bar{I}^{B}\left(T\right)=\exp\left(\frac{1}{2}\int_{0}^{T}L_{f}^{B}\left(t\right)dt\right) \!$, $L_{f}^{B}(t)$ and $L_{\boldsymbol{S}_{\boldsymbol{\theta}}}^{B}\left(t\right)$ mean the continuous Lipschitz constant shown in \textit{Appendix B}.
\end{theorem}
Theorem \ref{theorem:Wasserstein_distance_upper_bound_DM} states that for DDIB-based methods whenever $\mathcal{W}_{2}(p_{T}^{A},p_{T}^{B})\ne0$ holds, $\mathcal{W}_{2}(p_{0}^{B},q_{0}^{B})\ne0$ must follow. To address this theoretical gap, we compute the OT map $M_{ot,T}^{A\to B}$ to align the source domain latent code distribution $p_{T}^{A}$ with the target domain latent code distribution $p_{T}^{B}$ (See Algorithm \ref{alg:Computing_OT}). We then use $M_{ot,T}^{A\to B}(p_{T}^{A})$ as the initial distribution for the reverse process of DM$^{B}$, effectively eliminating the discrepancy caused by latent code distribution mismatch. Furthermore, we provide an error estimation in Theorem \ref{corollary:Wasserstein_distance_upper_bound_OUR}. The proof can be found in \textit{Appendix D.2}.
\begin{theorem}\label{corollary:Wasserstein_distance_upper_bound_OUR}
The error upper bound between the distribution obtained by OT-ALD and the ground truth is 
    \begin{equation*}
        \begin{split}
            \mathcal{W}_{2}(p_{0}^{B},q_{0}^{B})\leq \sqrt{2T}(\mathcal{J}_{SM}^{B})^{\frac{1}{2}}+KI^{B}(T) \| u_{T}^{A\to B}-u_{\boldsymbol{h}} \|_{\infty}^{\frac{1}{2}}
        \end{split},
    \end{equation*}
where $K$ is a positive constant and $u_{T}^{A\to B}$ is the true Brenier function between $p_{T}^{A}$ and $p_{T}^{B}$.
\end{theorem}
It is worth noting that the noise scaling factor $\eta \in [0,1]$ in \eqref{eq:solution_DMs} determines whether the diffusion process is Markovian or non-Markovian by controlling the noise added during sampling \cite{song2020denoising}. When $\eta \to 1$, our model generates images with greater diversity. Conversely, as $\eta \to 0$, it produces images more quickly and with higher determinism (See Fig. \ref{fig:eta_T:a}).

\textbf{Complexity analysis.} In Algorithm 1, the complexity of determining the cell to which each point in set $\mathcal{X}_{T}^{A}$ belongs is $\mathcal{O}(|\mathcal{K}||\mathcal{I}|)$. Subsequently, the complexity of calculating the measure of a cell based on the points belonging to it is $\mathcal{O}(|\mathcal{K}||\mathcal{I}|)$. Assuming the number of iterations in Algorithm \ref{alg:Computing_OT} is $N$, the complexity introduced by the Adam algorithm is $\mathcal{O}(N|\mathcal{I}|)$. Therefore, the total complexity of computing OT is $\mathcal{O}((2|\mathcal{K}|+N)|\mathcal{I}|)$.
\begin{algorithm}[H]
    \caption{Computing OT between latent distributions.}
    \label{alg:Computing_OT}
    \textbf{Input}: Source images $ \mathcal{X}_{0}^{A}=\{\boldsymbol{x}_{0,k}^{A}\}_{k\in\mathcal{K}}\sim p_{0}^{A} $, target images $ \mathcal{X}_{0}^{B}=\{\boldsymbol{x}_{0,i}^{B}\}_{i\in\mathcal{I}}\sim p_{0}^{B} $, trained $\boldsymbol{S}_{\boldsymbol{\theta}}^{A}$ and $\boldsymbol{S}_{\boldsymbol{\theta}}^{B}$, diffusion termination time $T$.\\
    \textbf{Output}: OT map $ M_{ot,T}^{A\to B} $.
    \begin{algorithmic}[1]
        \STATE $\mathcal{X}_{T}^{A}\longleftarrow Solver_{\eta}(\mathcal{X}_{0}^{A},\boldsymbol{S}_{\boldsymbol{\theta}}^{A},0,T)$.\\
        \STATE $\mathcal{X}_{T}^{B}\longleftarrow Solver_{\eta}(\mathcal{X}_{0}^{B},\boldsymbol{S}_{\boldsymbol{\theta}}^{B},0,T)$.\\
        \STATE $\boldsymbol{h}\longleftarrow \boldsymbol{0}$.\\
        \REPEAT
        \STATE Calculate $\nabla E(\boldsymbol{h})=(w_{i}(\boldsymbol{h})-\nu_{i})^{T}$.\\
        \STATE $\nabla E(\boldsymbol{h})=\nabla E(\boldsymbol{h})-mean(\nabla E(\boldsymbol{h}))$.\\
        \STATE Update $\boldsymbol{h}$ by Adam algorithm with $\beta_{1}=0.9$, $\beta_{2}=0.5$.
        \UNTIL{Converge}
        \STATE $M_{ot,T}^{A\to B}\longleftarrow\nabla u_{\boldsymbol{h}}$, $u_{\boldsymbol{h}}(\cdot)=\max\limits_{i} \langle \cdot,\boldsymbol{x}_{T,i}^{B}\rangle +h_{i}$.
    \end{algorithmic}
    \textbf{Return}: $ M_{ot,T}^{A\to B} $.
\end{algorithm}
\begin{algorithm}[H]
    \caption{I2I translation process of OT-ALD.}
    \label{alg:translation}
    \textbf{Input}: Source image $ \boldsymbol{x}_{0}^{A}\sim p_{0}^{A} $, computed OT map $ M_{ot,T}^{A\to B} $, trained score networks $ \boldsymbol{S}_{\boldsymbol{\theta}}^{A}(\boldsymbol{x}^{A},t) $ and $ \boldsymbol{S}_{\boldsymbol{\theta}}^{B}(\boldsymbol{x}^{B},t) $, noise scaling factor $\eta$.\\
    \textbf{Output}: Target image $ \boldsymbol{x}_{0}^{B} $.
    
    \begin{algorithmic}[1]
        \STATE $\boldsymbol{x}_{T}^{A}\longleftarrow Solver_{\eta}(\boldsymbol{x}_{0}^{A},\boldsymbol{S}_{\boldsymbol{\theta}}^{A},0,T)$.
        \STATE $ \boldsymbol{x}_{T}^{B}\longleftarrow M_{ot,T}^{A\to B}(\boldsymbol{x}_{T}^{A}) $.
        \STATE $\boldsymbol{x}_{0}^{B}\longleftarrow Solver_{\eta}(\boldsymbol{x}_{T}^{B},\boldsymbol{S}_{\boldsymbol{\theta}}^{B},T,0)$.
    \end{algorithmic}
    \textbf{Return}: $ \boldsymbol{x}_{0}^{B} $.
\end{algorithm}

Therefore, the OT-ALD consists of three stages: (1) Compute the latent code $\boldsymbol{x}_{T}^{A}$ by applying the forward process of DM$^{A}$ to degrade the source image $\boldsymbol{x}_{0}^{A}$. (2) Obtain the latent code $\boldsymbol{x}_{T}^{B}$ in the target domain corresponding to $\boldsymbol{x}_{T}^{A}$ under OT map $M_{ot,T}^{A\to B}$. (3) Generate the target image $\boldsymbol{x}_{0}^{B}$ by applying the reverse process of DM$^{B}$ to reconstruct $\boldsymbol{x}_{T}^{B}$. The complete translation process is outlined in Algorithm \ref{alg:translation}. 

Building on its ability to eliminate latent code distribution mismatch, OT-ALD can be naturally extended to accelerated I2I translation by reducing the diffusion termination time $T$. Fig. \ref{fig:eta_T:b} highlights that OT-ALD with a smaller $T$ better preserves the details and structural integrity of the source image, while a larger $T$ leverages the strengths of DM to enhance the diversity of the translated image. Therefore, an appropriate value of $T$ must be selected to strike a balance between computational efficiency and output quality. 

Theoretically, OT-ALD satisfies cycle consistency, which we explain from two different perspectives: sample-level (Theorem \ref{theorem:Sample cyclic consistency}) and distribution-level (Theorem \ref{theorem:Distribution cyclic consistency}). Notably, sample-level cycle consistency holds only when $\eta=0$ in Algorithm \ref{alg:translation}. However, once stochasticity is introduced, this desirable property no longer holds. In contrast, distribution-level cycle consistency remains valid for $\forall\eta\in[0,1]$ under the control of FPE \eqref{eq:Forward_Fokker_Planck}.

\begin{theorem}[Sample cycle consistency]\label{theorem:Sample cyclic consistency}
    Disregarding the score matching error, the discretization error of the $Solver_{\eta}$ and the computational error of the OT map, $\eta=0$. For any given $T\geq 0$ and source image $\boldsymbol{x}_{0}^{A}$, we consider the following cyclic process
    \begin{equation*}
        \begin{matrix}
         \boldsymbol{x}_{T}^{A}=Solver_{\eta}(\boldsymbol{x}_{0}^{A},\boldsymbol{S}_{\boldsymbol{\theta}}^{A},0,T), \boldsymbol{x}_{T}^{B}=M_{ot,T}^{A\to B}(\boldsymbol{x}_{T}^{A}),\\ \boldsymbol{x}_{0}^{B}=Solver_{\eta}(\boldsymbol{x}_{T}^{B},\boldsymbol{S}_{\boldsymbol{\theta}}^{B},T,0),
         \boldsymbol{x}_{T}^{\prime B}=Solver_{\eta}(\boldsymbol{x}_{T}^{B},\boldsymbol{S}_{\boldsymbol{\theta}}^{B},0,T),\\ \boldsymbol{x}_{T}^{\prime A}=M_{ot,T}^{B\to A}(\boldsymbol{x}_{T}^{\prime B}), \boldsymbol{x}_{0}^{\prime A}=Solver_{\eta}(\boldsymbol{x}_{T}^{\prime A},\boldsymbol{S}_{\boldsymbol{\theta}}^{A},T,0),
        \end{matrix}
    \end{equation*}
    where $M_{ot,T}^{B\to A}=(M_{ot,T}^{A\to B})^{-1}$, then $\|\boldsymbol{x}_{0}^{A}-\boldsymbol{x}_{0}^{\prime A}\|_{2}=0$ holds. See \textit{Appendix D.3} for the proof detail.
\end{theorem}
\begin{theorem}[Distributional cycle consistency]\label{theorem:Distribution cyclic consistency}
    Under the same setting as Theorem \ref{theorem:Sample cyclic consistency}, we apply the proposed method to translate the source distribution $p_{0}^{A}$ into the target distribution. Subsequently, the approximate source distribution $p_{0}^{\prime A}$ derived from the reverse translation action on the target distribution. Then we have $\mathcal{W}_{2}(p_{0}^{A},p_{0}^{\prime A})=0$ for $\forall\eta\in[0,1]$. See \textit{Appendix D.4} for the proof detail.
\end{theorem}
Sample consistency forms the basis of distribution consistency—when all individuals satisfy cycle consistency, the distribution naturally aligns. However, distribution consistency does not require strict reversibility of individual samples, making it more flexible and tolerant of individual differences to improve overall distribution alignment.
\begin{figure} [H]
    \centering
    \renewcommand{\figurename}{Fig}
    \includegraphics[width=1\columnwidth]{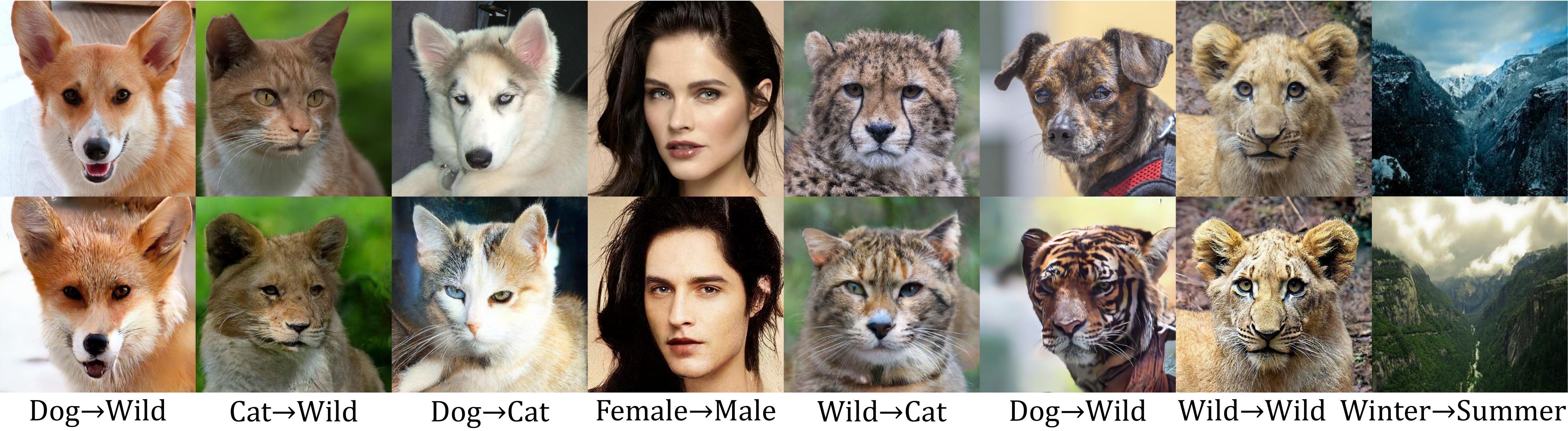}%
    \caption{The flexibility tests of OT-ALD on the trained DMs and OT map. Top-row images are sources; bottom-row images are the corresponding outputs generated by OT-ALD.}\label{Visual_contrast9}
\end{figure}
Moreover, OT-ALD is based on independently trained DMs and OT map, thereby inheriting the strong flexibility advantages of DDIB-based methods. To validate this property, we recombine the trained DMs in a manner different from the four main experiments. As shown in Fig. \ref{Visual_contrast9}, OT-ALD still achieves high-quality translation results under this setting. Notably, as shown in the "Wild→Wild" results in Fig. \ref{Visual_contrast9}, when the target image dataset is the same as the source image dataset, our model can achieve the effect of detail enhancement.

\section{Experiments}
\subsection{Implementation details}
\textbf{Datasets.} We evaluate four I2I translation tasks on three public datasets. (1) The AFHQ dataset \cite{choi2020stargan} is a 512$\times$512 animal face dataset containing approximately 5,000 images per class (cat, dog, wild), used for Cat$\to$Dog and Wild$\to$Dog translation. (2) The Summer2Winter dataset \cite{zhu2017unpaired} consists of 256$\times$256 seasonal landscape images under varying weather and lighting conditions, used for Summer$\to$Winter translation. (3) The CelebA-HQ dataset \cite{karras2017progressive} includes 30,000 facial images (resized to 512$\times$512) with diverse attributes and poses, used for Male$\to$Female translation.

\textbf{Baseline models.} In this work, we compare our model with the following methods: CUT \cite{park2020contrastive}, EGSDE \cite{zhao2022egsde}, ILVR \cite{choi2021ilvr}, SDEdit \cite{meng2021sdedit}, StarGANv2 \cite{choi2020stargan}, SDDM \cite{sun2023sddm}, InjectDiff \cite{li2023injecting}, CycleGAN \cite{zhu2017unpaired}, QS-Attn \cite{hu2022qs}, DDIB \cite{su2022dual},
InstPix2Pix \cite{brooks2023instructpix2pix}, Pix2Pix-Zero \cite{parmar2023zero}, P2P \cite{hertz2022prompt}, 
UNSB \cite{kim2023unpaired}, 
GcGAN \cite{fu2019geometry}, DistanceGAN \cite{benaim2017one}, Plug\&Play \cite{tumanyan2023plug}, CycleDiff \cite{wu2023latent}, Turbo \cite{parmar2024one}, 
U-GAT-IT \cite{kim2019u}, NICE-GAN \cite{chen2020reusing}, DDIB \cite{su2022dual,yu2023cross,popov2023optimal,zhang2024decdm,bourou2024phendiff,hur2024expanding,yin2024scalable,mancusi2024latent}, DMT \cite{xia2024diffusion}.

\textbf{Evaluation metrics.} For the translated images, we evaluate realism using Fréchet Inception Distance (FID) \cite{heusel2017gans}, Kernel Inception Distance (KID) \cite{binkowski2018demystifying} and Fool rates \cite{saharia2022image}. To measure faithfulness, we employ $L_{2}$ distance, Peak Signal-to-Noise Ratio (PSNR), Structural Similarity Index Measure (SSIM) \cite{wang2004image}, and DINO-Struct-Dist \cite{tumanyan2022splicing}. Additionally, the Wasserstein distance $\mathcal{W}_{2}(p_{T}^{B},q_{T}^{B})$ quantifies the latent code distribution matching error of the proposed method and DDIB-based approach. Finally, we assess computational efficiency by comparing translation time (s/image) across different models.

More experimental configuration details can be found in \textit{Appendix E.1}.
\subsection{Mismatch within latent code distributions and cycle consistency}
\begin{figure} [htbp]
    \centering
    \renewcommand{\figurename}{Fig}
    
    \subfigure[Sample cycle consistency\label{fig:Cyclic_consistency:a}]{\includegraphics[width=0.8\columnwidth]{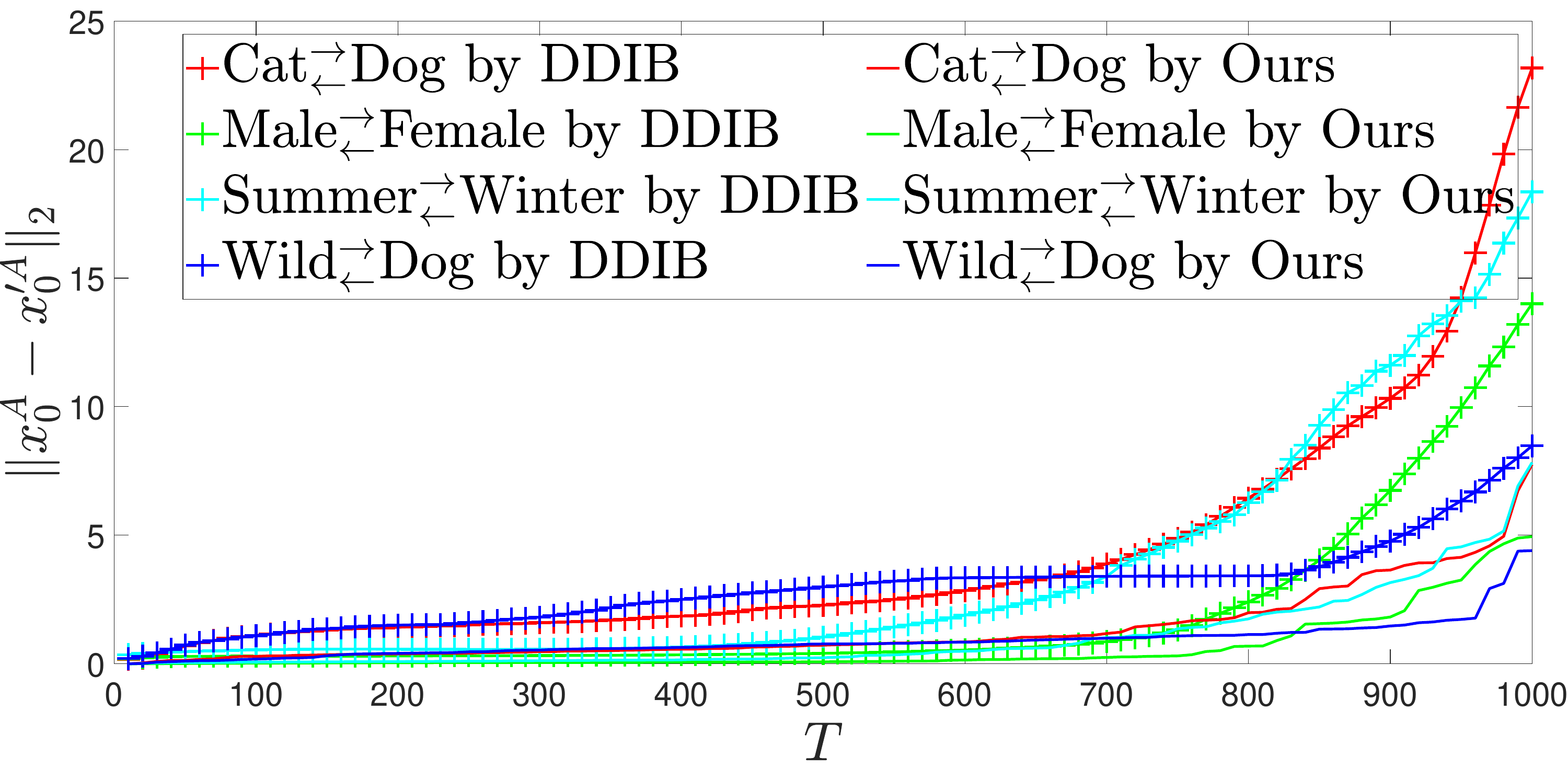}}
    \subfigure[Distribution cycle consistency\label{fig:Cyclic_consistency:b}]{\includegraphics[width=0.8\columnwidth]{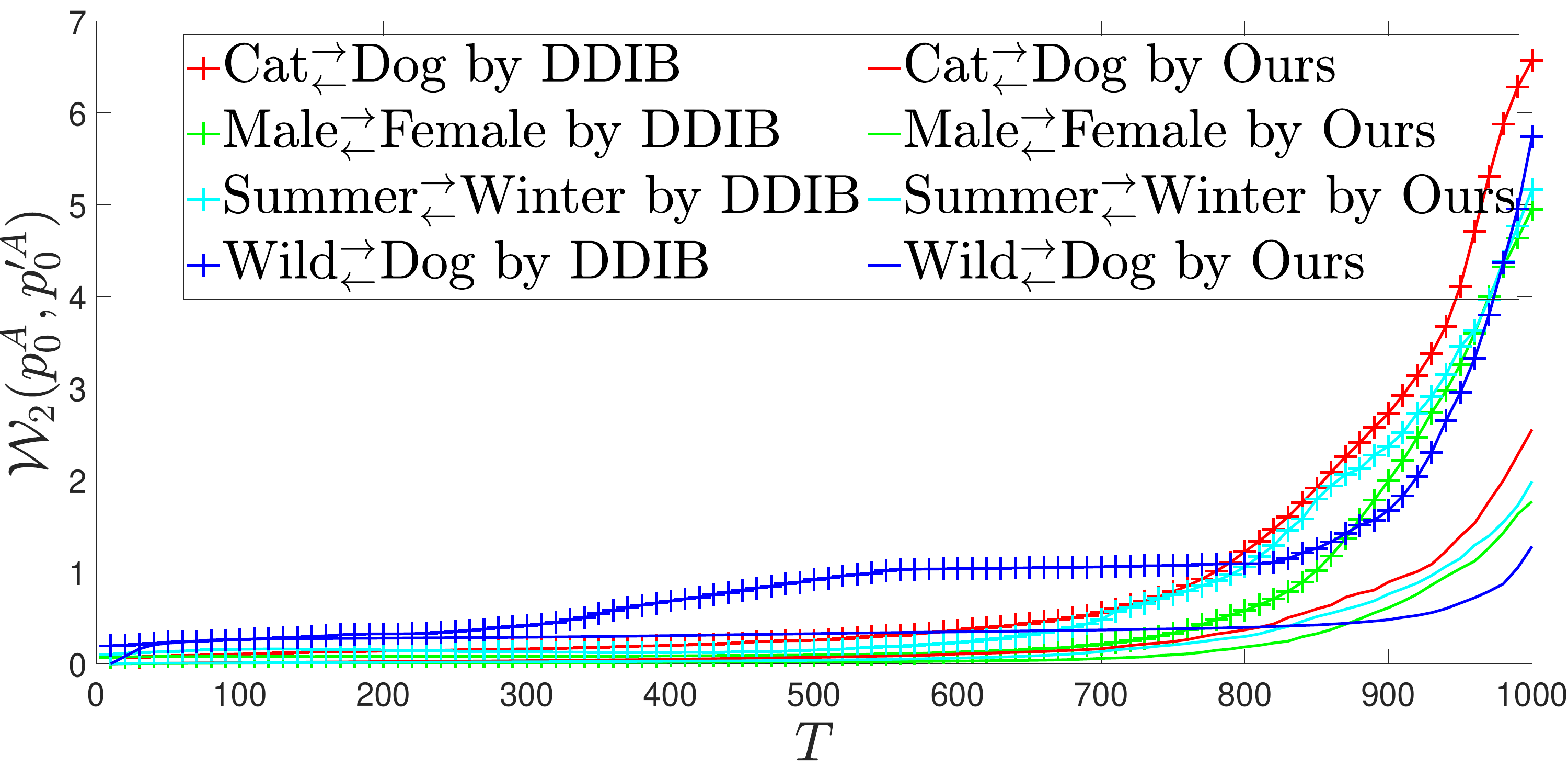}}
    \caption{Comparison of cycle consistency between OT-ALD and DDIB-based methods.}\label{Cyclic_consistency}
\end{figure}

For the four I2I translation tasks, Fig. \ref{fig:Toy_example:b} shows the latent distribution mismatch for our method and the DDIB-based approach. The results confirm that introducing the OT map effectively eliminates this discrepancy, yielding a target distribution closer to the ground truth (see Fig. \ref{fig:Toy_example:c}, Theorems \ref{theorem:Wasserstein_distance_upper_bound_DM}–\ref{corollary:Wasserstein_distance_upper_bound_OUR}). We further analyze the effect of varying diffusion termination time $T$, showing that our method is less sensitive to $T$ than DDIB, which requires a longer diffusion process to compensate for the mismatch—at the cost of significantly lower sampling efficiency.

Subsequently, we evaluate the cycle consistency of our method and DDIB-based approach in Figs. \ref{fig:Cyclic_consistency:a} and \ref{fig:Cyclic_consistency:b}. Although both methods theoretically satisfy sample-level and distribution-level cycle consistency (Theorems \ref{theorem:Sample cyclic consistency}-\ref{theorem:Distribution cyclic consistency}), the large number of diffusion steps in the DDIB-based method leads to accumulated computational errors, resulting in inferior cycle consistency compared to our model.
\begin{table}[hbtp]
\caption{Quantitative comparison on four tasks. $\uparrow$ and $\downarrow$ indicate directions of better values, where the optimal value is marked in \textbf{bold} and the suboptimal is marked with \underline{underline}.}\label{quantitative results}
\centering
\scalebox{0.88}{
\begin{tabular}{l|cccc}
\hline
\multirow{2}{*}{Methods} & \multicolumn{4}{c}{Cat$\to$Dog} \\ \cline{2-5} 
& FID$\downarrow$  & $L_{2}\downarrow$ & PSNR$\uparrow$ & SSIM$\uparrow$ \\ \hline
CUT & 76.21 & 59.78 & 17.48 & \textbf{0.601} \\
StarGANv2 & 54.88 & 133.65 & 10.63 & 0.27\\
CycleGAN & 85.90 & --- & 18.13 & \underline{0.58} \\
QS-Attn & 72.80 & --- & 18.26 & 0.39 \\
\hline
SDEdit & 74.14 & 47.88 & 19.19 & 0.423 \\
SDDM & 62.29 & --- & --- & 0.422 \\
SDDM$^{\dagger}$ & 49.43 & --- & --- & 0.361 \\
InjectDiff & 63.76 & --- & 20.10 & 0.48 \\
ILVR & 74.37 & 56.95 & 17.77 & 0.363 \\
EGSDE & 65.82 & 47.22 & 19.31 & 0.415 \\
EGSDE$^{\dagger}$ & 51.04 & 62.02 & 17.17 & 0.361 \\
DMT & \underline{47.86} & \underline{45.98} & \underline{20.55} & 0.412 \\
DDIB & 59.62 & 60.23 & 18.91 & 0.372 \\
\hline
OT-ALD & \textbf{44.31} & \textbf{42.35} & \textbf{21.20} & 0.470\\ \hline
\hline
\multirow{2}{*}{Methods} & \multicolumn{4}{c}{Wild$\to$Dog} \\ \cline{2-5} 
& FID$\downarrow$  & $L_{2}\downarrow$ & PSNR$\uparrow$ & SSIM$\uparrow$ \\ \hline
CUT & 92.94 & 62.21 & 17.20 & \textbf{0.592} \\
ILVR & 75.33 & 63.40 & 16.85 & 0.287 \\
SDEdit & 68.51 & 55.36 & 17.98 & 0.343 \\
EGSDE & 59.75 & 54.34 & \underline{18.14} & 0.343 \\
EGSDE$^{\dagger}$ & 50.43 & 66.52 & 16.40 & 0.300 \\
SDDM$^{\dagger}$ & 57.38 & --- & --- & 0.328 \\
DMT & \underline{49.83} & \underline{52.42} & 18.05 & 0.338 \\
DDIB & 55.34 & 60.01 & 17.72 & 0.332 \\
\hline
OT-ALD & \textbf{47.65} & \textbf{50.03} & \textbf{18.95} & \underline{0.351}\\ \hline
\hline
\multirow{2}{*}{Methods} & \multicolumn{4}{c}{Male$\to$Female} \\ \cline{2-5} 
& FID$\downarrow$  & $L_{2}\downarrow$ & PSNR$\uparrow$ & SSIM$\uparrow$ \\ \hline
CUT & 31.94 & 46.61 & 19.87 & \underline{0.74} \\
ILVR & 46.12 & 52.17 & 18.59 & 0.510 \\
SDEdit & 49.43 & 43.70 & 20.03 & 0.572 \\
EGSDE & 41.93 & 42.04 & 20.35 & 0.574 \\
EGSDE$^{\dagger}$ & 30.61 & 53.44 & 18.32 & 0.510 \\
InjectDiff & \underline{28.12} & --- & 21.65 & 0.67 \\
CycleGAN & 36.75 & --- & 21.54 & 0.70 \\
QS-Attn & 32.56 & --- & 20.68 & 0.60 \\
SDDM & 44.37 & --- & --- & 0.526 \\
DMT & 29.01 & \underline{35.63} & \underline{22.42} & 0.687 \\
DDIB & 38.25 & 38.55 & 21.37 & 0.66 \\
\hline
OT-ALD & \textbf{25.21} & \textbf{31.51} & \textbf{23.05} & \textbf{0.75}\\ 
\end{tabular}}
\scalebox{0.92}{
\begin{tabular}{l|cccc}
\hline
\hline
\multirow{2}{*}{Methods} & \multicolumn{3}{c}{Summer$\to$Winter} \\ \cline{2-4} 
& FID$\downarrow$  & KID$\downarrow$ & DINO Struct.$\downarrow$ \\ \hline
CycleGAN & 62.9 & 1.022 & 2.6 \\
DistanceGAN & 97.2 & 2.843 & --- \\
GcGAN & 97.5 & 2.755 & --- \\
CUT & 72.1 & 1.207 & 2.1 \\
SDEdit & 66.1 & 3.218 & 2.1 \\
Plug\&Play & 67.3 & --- & 2.8 \\
CycleDiff & 64.1 & --- & 3.6 \\
P2P & 99.1 & 2.626 & --- \\
Pix2Pix-Zero & 68.0 & --- & 3.0 \\
InstPix2Pix & 68.3 & --- & 3.7 \\
UNSB & 73.9 & \textbf{0.421} & --- \\
DDIB & 90.8 & 2.36 & 7.2 \\
Turbo & 56.3 & --- & \textbf{0.6} \\
NiceGAN & 76.03 & 0.67 & --- \\
NiceGAN$^{\dagger}$ & 77.13 & 0.73 & --- \\
U-GAT-IT & 88.41 & 1.43 & --- \\
EGSDE & 62.38 & 0.75 & 1.8 \\
ILVR & 65.64 & 1.35 & 2.0 \\
DMT & \underline{54.64} & 0.564 & 1.7 \\
\hline
OT-ALD & \textbf{52.87} & \underline{0.521} & \underline{1.5}\\
\hline
\end{tabular}}
\end{table}
\subsection{Image realism and faithfulness}
Tab. \ref{quantitative results} reports the quantitative results of OT-ALD ($T=500$, $\eta=0.2$) and baseline models across four I2I translation tasks. Compared to other methods, OT-ALD not only preserves the structural integrity of the source images but also achieves the highest output quality.
\begin{figure} [htbp]
    \centering
    \renewcommand{\figurename}{Fig}
    \includegraphics[width=1\columnwidth]{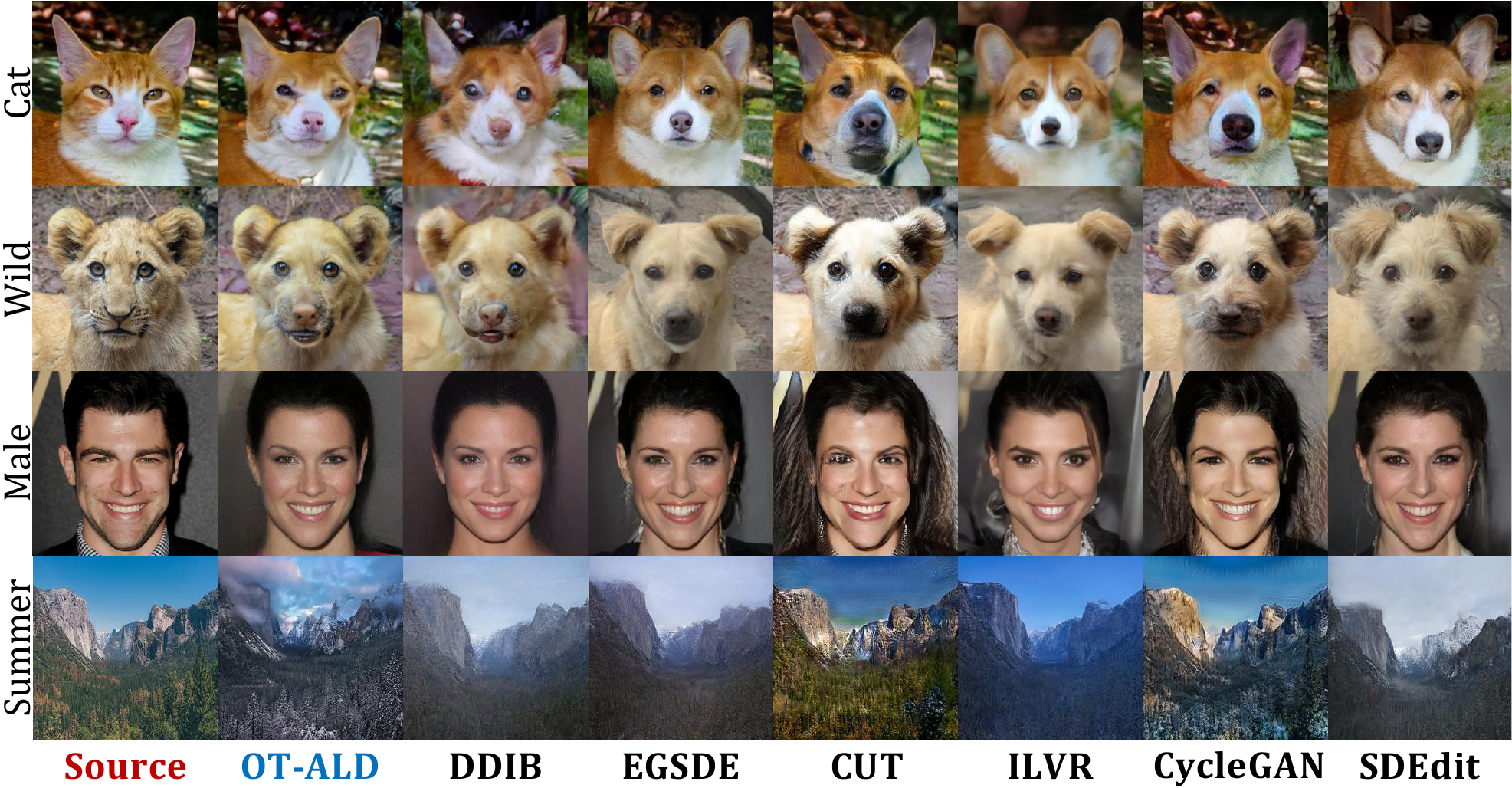}%
    \caption{Visualization comparison of different models over four I2I translation tasks. Compared with other mainstream methods, OT-ALD has the highest fidelity to source images, and its output quality is on par with or even superior to them.}\label{Visual_contrast}
\end{figure}
Across these four tasks, the FID value of our model is 2.6 lower on average than that of the optimal model in each corresponding task. Furthermore, Fig. \ref{Visual_contrast} provides a visual comparison between OT-ALD and representative I2I approaches, demonstrating that our model generates clearer and more realistic results than its counterparts, while largely maintaining the structure of the source image.

Human evaluations offer a more accurate assessment of I2I translation quality. As shown in \cite{mantiuk2012comparison}, forced pairwise comparison is a reliable method for image quality evaluation. Accordingly, we adopt the 2-alternative forced-choice (2AFC) paradigm \cite{zhang2016colorful}, where 50 participants are shown 20 randomly selected image pairs, each consisting of a generated image (from the source domain) and its corresponding real target image. Participants are informed that only one image per pair is real and are asked to choose the one they believe is real. The proportion of a generated image is mistakenly selected—Fool Rate—is recorded. The Fool Rate value closer to 50\% indicates higher image realism, while a lower rate suggests more noticeable artifacts.
\begin{figure} [htbp]
    \centering
    \renewcommand{\figurename}{Fig}
    \includegraphics[width=1\columnwidth]{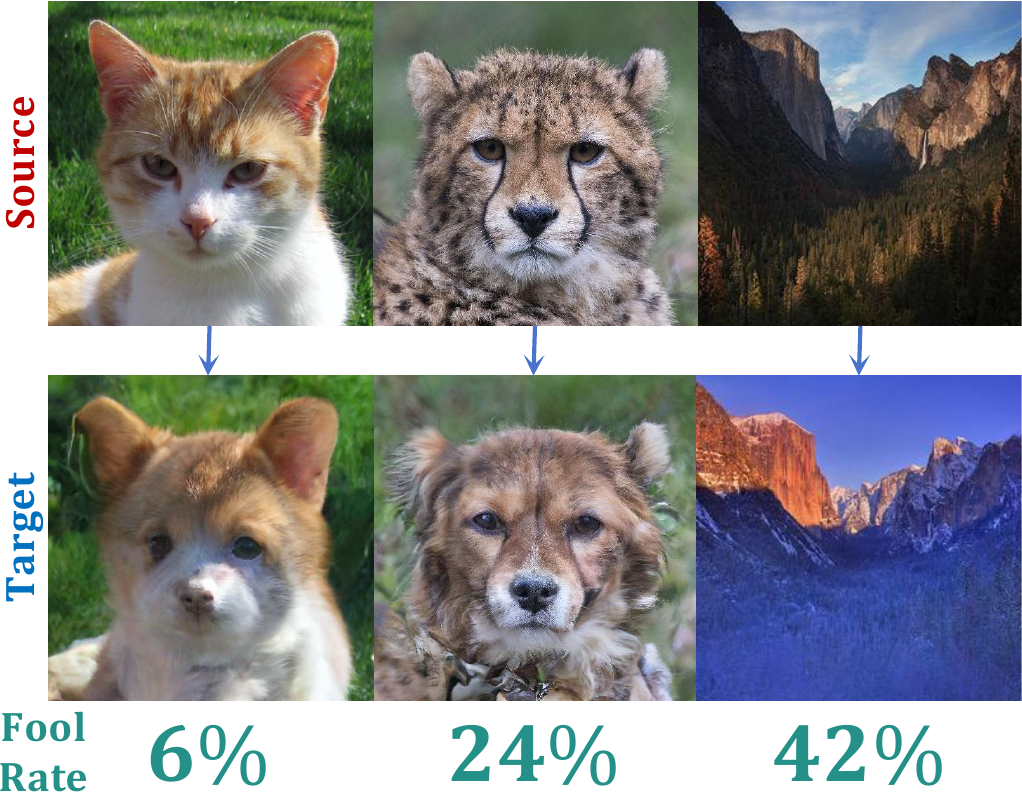}%
    \caption{Visualization comparison of Fool Rate. The Fool rate essentially reflects the difficulty in distinguishing between translated images and real images. The larger the value of this metric and the closer it is to 50\%, the stronger the realism of the images.}\label{fig:Fool Rate}
\end{figure}
\begin{table}[htbp]
\caption{Comparison of average Fool Rates ($\uparrow$) of different methods.}\label{table:Fool rates}
\centering
\scalebox{0.73}{
\begin{tabular}{l|cccc}
\hline
Methods & Cat$\to$Dog & Wild$\to$Dog & Summer$\to$Winter & Male$\to$Female\\ \hline
CUT      & \underline{23.6\%} & 22.3\% & 29.9\% & 28.5\% \\
CycleGAN & 22.3\% & 25.4\% & \underline{34.5\%} & 23.7\% \\
ILVR     & 21.4\% & 24.1\% & 23.3\% & \underline{29.2\%} \\
EGSDE    & 24.1\% & \textbf{28.3\%} & 26.6\% & 27.2\% \\
DDIB     & 18.5\% & 19.9\% & 30.8\% & 25.9\% \\
SDEdit   & 20.7\% & 21.5\% & 33.4\% & 24.4\% \\
DMT   & \textbf{24.3\%} & 25.7\% & 33.9\% & 27.5\% \\ \hline
OT-ALD     & 22.8\% & \underline{26.9\%} & \textbf{37.2\%} & \textbf{29.4\%} \\ \hline
\end{tabular}}
\end{table}

Fig. \ref{fig:Fool Rate} shows examples generated by OT-ALD with their corresponding average Fool Rates. Some results from the summer-to-winter task achieve rates above 40\%, indicating high realism, while some distorted images score near 0\%. Tab. \ref{table:Fool rates} compares the average Fool Rates of our model with others across four tasks, where our model consistently achieves around 29\%, suggesting it often produces images that are partially indistinguishable from real ones.
\subsection{Translation efficiency}
To evaluate the computational efficiency of different models on various translation tasks, we first fully train them on a single RTX 4090 GPU. Then, for each task, we randomly select 1000 source images and measure the average GPU time required by each model to translate them into target images. The results in Tab. \ref{tab:Translating time} demonstrate that OT-ALD is more efficient than other diffusion-based I2I translation technologies under same conditions. This advantage mainly stems from the reduced $T$ required by our model.

\begin{table}[htbp]
\caption{Translating time comparison (s/image, $\downarrow$) of different \textit{diffusion-based} models \textit{under same conditions}. The corresponding FID scores ($\downarrow$) are indicated in parentheses. Our model adopts $T=500$ and $\eta=0.2$. The average sampling time of our model across four tasks is 20.29\% lower than that of the model with the best translation efficiency (ILVR).}\label{tab:Translating time}
\centering
\scalebox{0.72}{
\begin{tabular}{l|cccc}
\hline
Methods & Cat$\to$Dog & Wild$\to$Dog & Summer$\to$Winter & Male$\to$Female \\ \hline
EGSDE   & 79.85 (51.04) & 79.75 (40.43) & 19.06 (62.38) & 79.80 (30.61) \\
ILVR    & 52.31 (70.37)  & 52.26 (75.33) & 16.97 (65.64) &  52.58 (46.12) \\
DDIB    &  92.30 (59.62) & 92.36 (55.34) & 20.64 (90.80) & 92.35 (38.25) \\
DMT    &  75.20 (47.86) & 74.98 (49.83) & 18.85 (54.64) & 75.25 (29.01) \\
SDEdit  &  86.44 (74.14) &  87.12 (68.51) &  19.85 (66.10) &  86.77 (49.43) \\
\hline
OT-ALD & \textbf{45.15 (44.31)} & \textbf{45.18 (47.65)} & \textbf{10.21 (52.87)} & \textbf{45.18 (25.21)} \\ \hline
\end{tabular}}
\end{table}
In addition to keeping the experimental equipment consistent, when testing the translating time in Tab. \ref{tab:Translating time}, we adopted the same sampling method for all diffusion-based models and did not use any acceleration strategies for the diffusion process. This means that if an acceleration strategy is introduced, all time data in Tab. \ref{tab:Translating time} will decrease synchronously, while our model will still outperform other baseline models.
\subsection{Ablation study}\label{sec:Ablation study}
(1) \textbf{Optimal Transport.} Figs. \ref{fig:Toy_example:b}-\ref{fig:Toy_example:c} and Fig. \ref{Cyclic_consistency} compare the classical DDIB-based method and OT-ALD in terms of latent distribution alignment, translated target distribution, and cycle consistency, along with their variations over diffusion termination time $T$.

(2) \textbf{Noise Scale Factor \(\eta\).} Fig. \ref{fig:eta_T:a} illustrates the impact of \(\eta\) on OT-ALD. With fixed \( T \), a larger \(\eta\) leads to improved output quality and diversity.

(3) \textbf{Diffusion termination time \( T \).}  Fig. \ref{fig:eta_T:b} demonstrates the effect of \( T \) on OT-ALD. When \(\eta\) is fixed and positive, increasing \( T \) enhances both generation quality and diversity. Moreover, for fixed-size image tasks, the average translation time of OT-ALD increases linearly with \( T \).
\section{Conclusion}
This paper provides a theoretical analysis of the limitations of classical DDIB-based I2I translation methods, namely: (1) low efficiency during the translation process and (2) translation trajectory deviations due to mismatched latent distributions. To address these issues, we propose OT-ALD, which effectively eliminates latent distribution mismatch, and offers theoretical guarantees for key properties such as cycle consistency. In practice, OT-ALD enables fast, high-quality image translation with flexible control over transformation objectives, allowing users to balance generation speed and synthesis quality according to their specific needs.
\section{Acknowledgments}
This research was supported by the National Key Research and Development Program of China under Grant No. 2021YFA1003003; the National Natural Science Foundation of China under Grant No. T2225012.
\clearpage
\newpage

\bibliography{aaai2026}

@article{lei2020geometric,
  title={A geometric understanding of deep learning},
  author={Lei, Na and An, Dongsheng and Guo, Yang and Su, Kehua and Liu, Shixia and Luo, Zhongxuan and Yau, Shing-Tung and Gu, Xianfeng},
  journal={Engineering},
  volume={6},
  pages={361--374},
  year={2020},
  publisher={Elsevier}
}

@inproceedings{an2020ae,
  title={AE-OT: A New Generative Model Based on Extended Semi-Discrete Optimal Transport},
  author={An, Dongsheng and Guo, Yang and Lei, Na and Luo, Zhongxuan and Yau, Shing-Tung and Gu, Xianfeng},
  booktitle={Proceedings of the 8th International Conference on Learning Representations},
  year={2020}
}

@article{brenier1991polar,
  title={Polar factorization and monotone rearrangement of vector-valued functions},
  author={Brenier, Yann},
  journal={Communications on pure and applied mathematics},
  volume={44},
  pages={375--417},
  year={1991},
  publisher={Wiley Online Library}
}

@inproceedings{ho2020denoising,
  title={Denoising diffusion probabilistic models},
  author={Ho, Jonathan and Jain, Ajay and Abbeel, Pieter},
  booktitle={Proceedings of the 34th International Conference on Neural Information Processing Systems},
  pages={6840--6851},
  year={2020}
}

@inproceedings{khrulkov2022understanding,
  title={Understanding DDPM Latent Codes Through Optimal Transport},
  author={Khrulkov, Valentin and Ryzhakov, Gleb and Chertkov, Andrei and Oseledets, Ivan},
  booktitle={The Eleventh International Conference on Learning Representations},
  year={2022}
}

@article{song2021maximum,
title={Maximum likelihood training of score-based diffusion models},
author={Song, Yang and Durkan, Conor and Murray, Iain and Ermon, Stefano},
journal={Advances in neural information processing systems},
volume={34},
pages={1415--1428},
year={2021}
}

@book{risken1996fokker,
title={Fokker-planck equation},
author={Risken, Hannes and Risken, Hannes},
year={1996},
publisher={Springer}
}

@article{song2020score,
title={Score-based generative modeling through stochastic differential equations},
author={Song, Yang and Sohl-Dickstein, Jascha and Kingma, Diederik P and Kumar, Abhishek and Ermon, Stefano and Poole, Ben},
journal={arXiv preprint arXiv:2011.13456},
year={2020}
}

@article{anderson1982reverse,
title={Reverse-time diffusion equation models},
author={Anderson, Brian DO},
journal={Stochastic Processes and their Applications},
volume={12},
number={3},
pages={313--326},
year={1982},
publisher={Elsevier}
}

@article{gu2013variational,
title={Variational principles for Minkowski type problems, discrete optimal transport, and discrete Monge-Ampere equations},
author={Gu, Xianfeng and Luo, Feng and Sun, Jian and Yau, S-T},
journal={arXiv preprint arXiv:1302.5472},
year={2013}
}

@article{kwon2022score,
title={Score-based generative modeling secretly minimizes the wasserstein distance},
author={Kwon, Dohyun and Fan, Ying and Lee, Kangwook},
journal={Advances in Neural Information Processing Systems},
volume={35},
pages={20205--20217},
year={2022}
}

@article{franzese2023much,
title={How much is enough? a study on diffusion times in score-based generative models},
author={Franzese, Giulio and Rossi, Simone and Yang, Lixuan and Finamore, Alessandro and Rossi, Dario and Filippone, Maurizio and Michiardi, Pietro},
journal={Entropy},
volume={25},
number={4},
pages={633},
year={2023},
publisher={MDPI}
}

@article{su2022dual,
title={Dual diffusion implicit bridges for image-to-image translation},
author={Su, Xuan and Song, Jiaming and Meng, Chenlin and Ermon, Stefano},
journal={arXiv preprint arXiv:2203.08382},
year={2022}
}

@inproceedings{merigot2020quantitative,
title={Quantitative stability of optimal transport maps and linearization of the 2-Wasserstein space},
author={M{\'e}rigot, Quentin and Delalande, Alex and Chazal, Frederic},
booktitle={International Conference on Artificial Intelligence and Statistics},
pages={3186--3196},
year={2020},
organization={PMLR}
}

@article{carrillo2006contractions,
title={Contractions in the 2-Wasserstein length space and thermalization of granular media},
author={Carrillo, Jos{\'e} A and McCann, Robert J and Villani, C{\'e}dric},
journal={Archive for Rational Mechanics and Analysis},
volume={179},
pages={217--263},
year={2006},
publisher={Springer}
}

@inproceedings{heusel2017gans,
  title={GANs trained by a two time-scale update rule converge to a local nash equilibrium},
  author={Heusel, Martin and Ramsauer, Hubert and Unterthiner, Thomas and Nessler, Bernhard and Hochreiter, Sepp},
  booktitle={Proceedings of the 31st International Conference on Neural Information Processing Systems},
  pages={6629--6640},
  year={2017}
}

@inproceedings{song2020denoising,
  title     = {Denoising Diffusion Implicit Models},
  author    = {Jiaming Song and Chenlin Meng and Stefano Ermon},
  booktitle = {International Conference on Learning Representations},
  year      = {2021},
}

@article{sasaki2021unit,
  title={Unit-ddpm: Unpaired image translation with denoising diffusion probabilistic models},
  author={Sasaki, Hiroshi and Willcocks, Chris G and Breckon, Toby P},
  journal={arXiv preprint arXiv:2104.05358},
  year={2021}
}

@article{zhao2022egsde,
  title={Egsde: Unpaired image-to-image translation via energy-guided stochastic differential equations},
  author={Zhao, Min and Bao, Fan and Li, Chongxuan and Zhu, Jun},
  journal={Advances in Neural Information Processing Systems},
  volume={35},
  pages={3609--3623},
  year={2022}
}

@inproceedings{saharia2022palette,
  title={Palette: Image-to-image diffusion models},
  author={Saharia, Chitwan and Chan, William and Chang, Huiwen and Lee, Chris and Ho, Jonathan and Salimans, Tim and Fleet, David and Norouzi, Mohammad},
  booktitle={ACM SIGGRAPH 2022 conference proceedings},
  pages={1--10},
  year={2022}
}

@article{wang2022pretraining,
  title={Pretraining is all you need for image-to-image translation},
  author={Wang, Tengfei and Zhang, Ting and Zhang, Bo and Ouyang, Hao and Chen, Dong and Chen, Qifeng and Wen, Fang},
  journal={arXiv preprint arXiv:2205.12952},
  year={2022}
}

@article{wang2022semantic,
  title={Semantic image synthesis via diffusion models},
  author={Wang, Weilun and Bao, Jianmin and Zhou, Wengang and Chen, Dongdong and Chen, Dong and Yuan, Lu and Li, Houqiang},
  journal={arXiv preprint arXiv:2207.00050},
  year={2022}
}

@inproceedings{li2023bbdm,
  title={Bbdm: Image-to-image translation with brownian bridge diffusion models},
  author={Li, Bo and Xue, Kaitao and Liu, Bin and Lai, Yu-Kun},
  booktitle={Proceedings of the IEEE/CVF conference on computer vision and pattern Recognition},
  pages={1952--1961},
  year={2023}
}

@article{kwon2022diffusion,
  title={Diffusion-based image translation using disentangled style and content representation},
  author={Kwon, Gihyun and Ye, Jong Chul},
  journal={arXiv preprint arXiv:2209.15264},
  year={2022}
}

@inproceedings{parmar2023zero,
  title={Zero-shot image-to-image translation},
  author={Parmar, Gaurav and Kumar Singh, Krishna and Zhang, Richard and Li, Yijun and Lu, Jingwan and Zhu, Jun-Yan},
  booktitle={ACM SIGGRAPH 2023 Conference Proceedings},
  pages={1--11},
  year={2023}
}

@article{meng2021sdedit,
  title={Sdedit: Guided image synthesis and editing with stochastic differential equations},
  author={Meng, Chenlin and He, Yutong and Song, Yang and Song, Jiaming and Wu, Jiajun and Zhu, Jun-Yan and Ermon, Stefano},
  journal={arXiv preprint arXiv:2108.01073},
  year={2021}
}

@inproceedings{zhu2017unpaired,
  title={Unpaired image-to-image translation using cycle-consistent adversarial networks},
  author={Zhu, Jun-Yan and Park, Taesung and Isola, Phillip and Efros, Alexei A},
  booktitle={Proceedings of the IEEE international conference on computer vision},
  pages={2223--2232},
  year={2017}
}

@inproceedings{isola2017image,
  title={Image-to-image translation with conditional adversarial networks},
  author={Isola, Phillip and Zhu, Jun-Yan and Zhou, Tinghui and Efros, Alexei A},
  booktitle={Proceedings of the IEEE conference on computer vision and pattern recognition},
  pages={1125--1134},
  year={2017}
}

@inproceedings{yi2017dualgan,
  title={Dualgan: Unsupervised dual learning for image-to-image translation},
  author={Yi, Zili and Zhang, Hao and Tan, Ping and Gong, Minglun},
  booktitle={Proceedings of the IEEE international conference on computer vision},
  pages={2849--2857},
  year={2017}
}

@inproceedings{kim2017learning,
  title={Learning to discover cross-domain relations with generative adversarial networks},
  author={Kim, Taeksoo and Cha, Moonsu and Kim, Hyunsoo and Lee, Jung Kwon and Kim, Jiwon},
  booktitle={International conference on machine learning},
  pages={1857--1865},
  year={2017},
  organization={PMLR}
}

@inproceedings{fu2019geometry,
  title={Geometry-consistent generative adversarial networks for one-sided unsupervised domain mapping},
  author={Fu, Huan and Gong, Mingming and Wang, Chaohui and Batmanghelich, Kayhan and Zhang, Kun and Tao, Dacheng},
  booktitle={Proceedings of the IEEE/CVF conference on computer vision and pattern recognition},
  pages={2427--2436},
  year={2019}
}

@inproceedings{park2020contrastive,
  title={Contrastive learning for unpaired image-to-image translation},
  author={Park, Taesung and Efros, Alexei A and Zhang, Richard and Zhu, Jun-Yan},
  booktitle={Computer Vision--ECCV 2020: 16th European Conference, Glasgow, UK, August 23--28, 2020, Proceedings, Part IX 16},
  pages={319--345},
  year={2020},
  organization={Springer}
}

@article{benaim2017one,
  title={One-sided unsupervised domain mapping},
  author={Benaim, Sagie and Wolf, Lior},
  journal={Advances in neural information processing systems},
  volume={30},
  year={2017}
}

@inproceedings{liang2021swinir,
  title={Swinir: Image restoration using swin transformer},
  author={Liang, Jingyun and Cao, Jiezhang and Sun, Guolei and Zhang, Kai and Van Gool, Luc and Timofte, Radu},
  booktitle={Proceedings of the IEEE/CVF international conference on computer vision},
  pages={1833--1844},
  year={2021}
}

@inproceedings{zamir2022restormer,
  title={Restormer: Efficient transformer for high-resolution image restoration},
  author={Zamir, Syed Waqas and Arora, Aditya and Khan, Salman and Hayat, Munawar and Khan, Fahad Shahbaz and Yang, Ming-Hsuan},
  booktitle={Proceedings of the IEEE/CVF conference on computer vision and pattern recognition},
  pages={5728--5739},
  year={2022}
}

@inproceedings{azadi2018multi,
  title={Multi-content gan for few-shot font style transfer},
  author={Azadi, Samaneh and Fisher, Matthew and Kim, Vladimir G and Wang, Zhaowen and Shechtman, Eli and Darrell, Trevor},
  booktitle={Proceedings of the IEEE conference on computer vision and pattern recognition},
  pages={7564--7573},
  year={2018}
}

@inproceedings{gatys2016image,
  title={Image style transfer using convolutional neural networks},
  author={Gatys, Leon A and Ecker, Alexander S and Bethge, Matthias},
  booktitle={Proceedings of the IEEE conference on computer vision and pattern recognition},
  pages={2414--2423},
  year={2016}
}

@inproceedings{rombach2022high,
  title={High-resolution image synthesis with latent diffusion models},
  author={Rombach, Robin and Blattmann, Andreas and Lorenz, Dominik and Esser, Patrick and Ommer, Bj{\"o}rn},
  booktitle={Proceedings of the IEEE/CVF conference on computer vision and pattern recognition},
  pages={10684--10695},
  year={2022}
}

@inproceedings{odena2017conditional,
  title={Conditional image synthesis with auxiliary classifier gans},
  author={Odena, Augustus and Olah, Christopher and Shlens, Jonathon},
  booktitle={International conference on machine learning},
  pages={2642--2651},
  year={2017},
  organization={PMLR}
}

@article{chazal2017inference,
  	title={Inference of curvature using tubular neighborhoods},
  	author={Chazal, Fr{\'e}d{\'e}ric and Cohen-Steiner, David and Lieutier, Andr{\'e} and M{\'e}rigot, Quentin and Thibert, Boris},
  	journal={Modern Approaches to Discrete Curvature},
  	pages={133--158},
  	year={2017},
  	publisher={Springer}
  }

@article{hoyez2022unsupervised,
  title={Unsupervised image-to-image translation: A review},
  author={Hoyez, Henri and Schockaert, C{\'e}dric and Rambach, Jason and Mirbach, Bruno and Stricker, Didier},
  journal={Sensors},
  volume={22},
  number={21},
  pages={8540},
  year={2022},
  publisher={MDPI}
}

@article{huang2024diffusion,
  title={Diffusion model-based image editing: A survey},
  author={Huang, Yi and Huang, Jiancheng and Liu, Yifan and Yan, Mingfu and Lv, Jiaxi and Liu, Jianzhuang and Xiong, Wei and Zhang, He and Chen, Shifeng and Cao, Liangliang},
  journal={arXiv preprint arXiv:2402.17525},
  year={2024}
}

@article{kim2024conditional,
  title={Conditional brownian bridge diffusion model for vhr sar to optical image translation},
  author={Kim, Seon-Hoon and Chung, Dae-won},
  journal={arXiv preprint arXiv:2408.07947},
  year={2024}
}

@article{kim2023unpaired,
  title={Unpaired Image-to-Image Translation via Neural Schr$\backslash$" odinger Bridge},
  author={Kim, Beomsu and Kwon, Gihyun and Kim, Kwanyoung and Ye, Jong Chul},
  journal={arXiv preprint arXiv:2305.15086},
  year={2023}
}

@article{tu2024unpaired,
  title={Unpaired image-to-image translation with diffusion adversarial network},
  author={Tu, Hangyao and Wang, Zheng and Zhao, Yanwei},
  journal={Mathematics},
  volume={12},
  number={20},
  pages={3178},
  year={2024},
  publisher={MDPI}
}

@article{luo2024target,
  title={Target-Guided Diffusion Models for Unpaired Cross-modality Medical Image Translation},
  author={Luo, Yimin and Yang, Qinyu and Liu, Ziyi and Shi, Zenglin and Huang, Weimin and Zheng, Guoyan and Cheng, Jun},
  journal={IEEE Journal of Biomedical and Health Informatics},
  year={2024},
  publisher={IEEE}
}

@inproceedings{kim2022diffusionclip,
  title={Diffusionclip: Text-guided diffusion models for robust image manipulation},
  author={Kim, Gwanghyun and Kwon, Taesung and Ye, Jong Chul},
  booktitle={Proceedings of the IEEE/CVF conference on computer vision and pattern recognition},
  pages={2426--2435},
  year={2022}
}

@inproceedings{tumanyan2023plug,
  title={Plug-and-play diffusion features for text-driven image-to-image translation},
  author={Tumanyan, Narek and Geyer, Michal and Bagon, Shai and Dekel, Tali},
  booktitle={Proceedings of the IEEE/CVF Conference on Computer Vision and Pattern Recognition},
  pages={1921--1930},
  year={2023}
}

@inproceedings{seo2023midms,
  title={Midms: Matching interleaved diffusion models for exemplar-based image translation},
  author={Seo, Junyoung and Lee, Gyuseong and Cho, Seokju and Lee, Jiyoung and Kim, Seungryong},
  booktitle={Proceedings of the AAAI Conference on Artificial Intelligence},
  volume={37},
  number={2},
  pages={2191--2199},
  year={2023}
}

@inproceedings{cheng2023general,
  title={General image-to-image translation with one-shot image guidance},
  author={Cheng, Bin and Liu, Zuhao and Peng, Yunbo and Lin, Yue},
  booktitle={Proceedings of the IEEE/CVF International Conference on Computer Vision},
  pages={22736--22746},
  year={2023}
}

@inproceedings{shi2024dragdiffusion,
  title={Dragdiffusion: Harnessing diffusion models for interactive point-based image editing},
  author={Shi, Yujun and Xue, Chuhui and Liew, Jun Hao and Pan, Jiachun and Yan, Hanshu and Zhang, Wenqing and Tan, Vincent YF and Bai, Song},
  booktitle={Proceedings of the IEEE/CVF Conference on Computer Vision and Pattern Recognition},
  pages={8839--8849},
  year={2024}
}

@inproceedings{yang2023paint,
  title={Paint by example: Exemplar-based image editing with diffusion models},
  author={Yang, Binxin and Gu, Shuyang and Zhang, Bo and Zhang, Ting and Chen, Xuejin and Sun, Xiaoyan and Chen, Dong and Wen, Fang},
  booktitle={Proceedings of the IEEE/CVF Conference on Computer Vision and Pattern Recognition},
  pages={18381--18391},
  year={2023}
}

@inproceedings{zhang2023sine,
  title={Sine: Single image editing with text-to-image diffusion models},
  author={Zhang, Zhixing and Han, Ligong and Ghosh, Arnab and Metaxas, Dimitris N and Ren, Jian},
  booktitle={Proceedings of the IEEE/CVF Conference on Computer Vision and Pattern Recognition},
  pages={6027--6037},
  year={2023}
}

@article{luo2023latent,
  title={Latent consistency models: Synthesizing high-resolution images with few-step inference},
  author={Luo, Simian and Tan, Yiqin and Huang, Longbo and Li, Jian and Zhao, Hang},
  journal={arXiv preprint arXiv:2310.04378},
  year={2023}
}

@article{jiang2024fast,
  title={Fast-DDPM: Fast Denoising Diffusion Probabilistic Models for Medical Image-to-Image Generation},
  author={Jiang, Hongxu and Imran, Muhammad and Ma, Linhai and Zhang, Teng and Zhou, Yuyin and Liang, Muxuan and Gong, Kuang and Shao, Wei},
  journal={arXiv e-prints},
  pages={arXiv--2405},
  year={2024}
}

@article{xia2024diffi2i,
  title={Diffi2i: efficient diffusion model for image-to-image translation},
  author={Xia, Bin and Zhang, Yulun and Wang, Shiyin and Wang, Yitong and Wu, Xinglong and Tian, Yapeng and Yang, Wenming and Timotfe, Radu and Van Gool, Luc},
  journal={IEEE Transactions on Pattern Analysis and Machine Intelligence},
  year={2024},
  publisher={IEEE}
}

@article{parmar2024one,
  title={One-step image translation with text-to-image models},
  author={Parmar, Gaurav and Park, Taesung and Narasimhan, Srinivasa and Zhu, Jun-Yan},
  journal={arXiv preprint arXiv:2403.12036},
  year={2024}
}

@article{xia2024diffusion,
  title={A diffusion model translator for efficient image-to-image translation},
  author={Xia, Mengfei and Zhou, Yu and Yi, Ran and Liu, Yong-Jin and Wang, Wenping},
  journal={IEEE Transactions on Pattern Analysis and Machine Intelligence},
  year={2024},
  publisher={IEEE}
}

@inproceedings{lee2024ebdm,
  title={EBDM: Exemplar-Guided Image Translation with Brownian-Bridge Diffusion Models},
  author={Lee, Eungbean and Jeong, Somi and Sohn, Kwanghoon},
  booktitle={European Conference on Computer Vision},
  pages={306--323},
  year={2024},
  organization={Springer}
}

@inproceedings{lee2024diffusion,
  title={Diffusion-Based Image-to-Image Translation by Noise Correction via Prompt Interpolation},
  author={Lee, Junsung and Kang, Minsoo and Han, Bohyung},
  booktitle={European Conference on Computer Vision},
  pages={289--304},
  year={2024},
  organization={Springer}
}

@article{wang2024mutual,
  title={Mutual information guided diffusion for zero-shot cross-modality medical image translation},
  author={Wang, Zihao and Yang, Yingyu and Chen, Yuzhou and Yuan, Tingting and Sermesant, Maxime and Delingette, Herv{\'e} and Wu, Ona},
  journal={IEEE Transactions on Medical Imaging},
  year={2024},
  publisher={IEEE}
}

@article{graf2023denoising,
  title={Denoising diffusion-based MRI to CT image translation enables automated spinal segmentation},
  author={Graf, Robert and Schmitt, Joachim and Schlaeger, Sarah and M{\"o}ller, Hendrik Kristian and Sideri-Lampretsa, Vasiliki and Sekuboyina, Anjany and Krieg, Sandro Manuel and Wiestler, Benedikt and Menze, Bjoern and Rueckert, Daniel and others},
  journal={European Radiology Experimental},
  volume={7},
  number={1},
  pages={70},
  year={2023},
  publisher={Springer}
}

@inproceedings{chencontourdiff,
  title={ContourDiff: Unpaired Image-to-Image Translation with Structural Consistency for Medical Imaging},
  author={Chen, Yuwen},
  booktitle={Medical Imaging with Deep Learning}
}

@article{li2023zero,
  title={Zero-shot medical image translation via frequency-guided diffusion models},
  author={Li, Yunxiang and Shao, Hua-Chieh and Liang, Xiao and Chen, Liyuan and Li, Ruiqi and Jiang, Steve and Wang, Jing and Zhang, You},
  journal={IEEE transactions on medical imaging},
  year={2023},
  publisher={IEEE}
}

@inproceedings{xing2024cross,
  title={Cross-conditioned diffusion model for medical image to image translation},
  author={Xing, Zhaohu and Yang, Sicheng and Chen, Sixiang and Ye, Tian and Yang, Yijun and Qin, Jing and Zhu, Lei},
  booktitle={International Conference on Medical Image Computing and Computer-Assisted Intervention},
  pages={201--211},
  year={2024},
  organization={Springer}
}

@article{patil2025mmit,
  title={MMIT-DDPM--Multilateral medical image translation with class and structure supervised diffusion-based model},
  author={Patil, Sanjeet S and Rajak, Rishav and Ramteke, Manojkumar and Rathore, Anurag S},
  journal={Computers in Biology and Medicine},
  volume={185},
  pages={109501},
  year={2025},
  publisher={Elsevier}
}

@article{guo2024learning,
  title={Learning SAR-to-optical image translation via diffusion models with color memory},
  author={Guo, Zhe and Liu, Jiayi and Cai, Qinglin and Zhang, Zhibo and Mei, Shaohui},
  journal={IEEE Journal of Selected Topics in Applied Earth Observations and Remote Sensing},
  year={2024},
  publisher={IEEE}
}

@article{zhang2025dcltv,
  title={DCLTV: An Improved Dual-Condition Diffusion Model for Laser-Visible Image Translation},
  author={Zhang, Xiaoyu and Zhang, Laixian and Guo, Huichao and Zheng, Haijing and Sun, Houpeng and Li, Yingchun and Li, Rong and Luan, Chenglong and Tong, Xiaoyun},
  journal={Sensors},
  volume={25},
  number={3},
  pages={697},
  year={2025},
  publisher={MDPI}
}

@article{vinholi2024multi,
  title={Multi-Sensor Diffusion-Driven Optical Image Translation for Large-Scale Applications},
  author={Vinholi, Jo{\~a}o Gabriel and Chini, Marco and Amziane, Anis and Machado, Renato and Silva, Danilo and Matgen, Patrick},
  journal={IEEE Journal of Selected Topics in Applied Earth Observations and Remote Sensing},
  year={2024},
  publisher={IEEE}
}

@article{bai2023conditional,
  title={Conditional Diffusion for SAR to Optical Image Translation},
  author={Bai, Xinyu and Pu, Xinyang and Xu, Feng},
  journal={IEEE Geoscience and Remote Sensing Letters},
  year={2023},
  publisher={IEEE}
}

@inproceedings{bai2024sar,
  title={Sar to optical image translation with color supervised diffusion model},
  author={Bai, Xinyu and Xu, Feng},
  booktitle={IGARSS 2024-2024 IEEE International Geoscience and Remote Sensing Symposium},
  pages={963--966},
  year={2024},
  organization={IEEE}
}

@article{shi2024brain,
  title={A brain-inspired approach for SAR-to-optical image translation based on diffusion models},
  author={Shi, Hao and Cui, Zihan and Chen, Liang and He, Jingfei and Yang, Jingyi},
  journal={Frontiers in Neuroscience},
  volume={18},
  pages={1352841},
  year={2024},
  publisher={Frontiers Media SA}
}

@inproceedings{ceylan2023pix2video,
  title={Pix2video: Video editing using image diffusion},
  author={Ceylan, Duygu and Huang, Chun-Hao P and Mitra, Niloy J},
  booktitle={Proceedings of the IEEE/CVF International Conference on Computer Vision},
  pages={23206--23217},
  year={2023}
}

@inproceedings{lugmayr2022repaint,
  title={Repaint: Inpainting using denoising diffusion probabilistic models},
  author={Lugmayr, Andreas and Danelljan, Martin and Romero, Andres and Yu, Fisher and Timofte, Radu and Van Gool, Luc},
  booktitle={Proceedings of the IEEE/CVF conference on computer vision and pattern recognition},
  pages={11461--11471},
  year={2022}
}

@inproceedings{li2023injecting,
  title={Injecting-Diffusion: Inject Domain-Independent Contents into Diffusion Models for Unpaired Image-to-Image Translation},
  author={Li, Luying and Ma, Lizhuang},
  booktitle={2023 IEEE International Conference on Multimedia and Expo (ICME)},
  pages={282--287},
  year={2023},
  organization={IEEE}
}

@inproceedings{zheng2023layoutdiffusion,
  title={Layoutdiffusion: Controllable diffusion model for layout-to-image generation},
  author={Zheng, Guangcong and Zhou, Xianpan and Li, Xuewei and Qi, Zhongang and Shan, Ying and Li, Xi},
  booktitle={Proceedings of the IEEE/CVF Conference on Computer Vision and Pattern Recognition},
  pages={22490--22499},
  year={2023}
}

@article{lin2023regeneration,
  title={Regeneration learning of diffusion models with rich prompts for zero-shot image translation},
  author={Lin, Yupei and Zhang, Sen and Yang, Xiaojun and Wang, Xiao and Shi, Yukai},
  journal={arXiv preprint arXiv:2305.04651},
  year={2023}
}

@article{han2024image,
  title={Image translation as diffusion visual programmers},
  author={Han, Cheng and Liang, James C and Wang, Qifan and Rabbani, Majid and Dianat, Sohail and Rao, Raghuveer and Wu, Ying Nian and Liu, Dongfang},
  journal={arXiv preprint arXiv:2401.09742},
  year={2024}
}

@article{lin2024mirrordiffusion,
  title={Mirrordiffusion: stabilizing diffusion process in zero-shot image translation by prompts redescription and beyond},
  author={Lin, Yupei and Xian, Xiaoyu and Shi, Yukai and Lin, Liang},
  journal={IEEE Signal Processing Letters},
  year={2024},
  publisher={IEEE}
}

@article{liu2024iidm,
  title={IIDM: Image-to-Image Diffusion Model for Semantic Image Synthesis},
  author={Liu, Feng and Chang, Xiaobin},
  journal={arXiv preprint arXiv:2403.13378},
  year={2024}
}

@inproceedings{gao2024frequency,
  title={Frequency-Controlled Diffusion Model for Versatile Text-Guided Image-to-Image Translation},
  author={Gao, Xiang and Xu, Zhengbo and Zhao, Junhan and Liu, Jiaying},
  booktitle={Proceedings of the AAAI Conference on Artificial Intelligence},
  volume={38},
  number={3},
  pages={1824--1832},
  year={2024}
}

@inproceedings{sun2023sddm,
  title={SDDM: score-decomposed diffusion models on manifolds for unpaired image-to-image translation},
  author={Sun, Shikun and Wei, Longhui and Xing, Junliang and Jia, Jia and Tian, Qi},
  booktitle={International Conference on Machine Learning},
  pages={33115--33134},
  year={2023},
  organization={PMLR}
}

@inproceedings{kim2024adaptive,
  title={Adaptive latent diffusion model for 3d medical image to image translation: Multi-modal magnetic resonance imaging study},
  author={Kim, Jonghun and Park, Hyunjin},
  booktitle={Proceedings of the IEEE/CVF Winter Conference on Applications of Computer Vision},
  pages={7604--7613},
  year={2024}
}

@article{yu2023cross,
  title={Cross: Diffusion model makes controllable, robust and secure image steganography},
  author={Yu, Jiwen and Zhang, Xuanyu and Xu, Youmin and Zhang, Jian},
  journal={Advances in neural information processing systems},
  volume={36},
  pages={80730--80743},
  year={2023}
}

@article{wang2025sindiffusion,
  title={Sindiffusion: Learning a diffusion model from a single natural image},
  author={Wang, Weilun and Bao, Jianmin and Zhou, Wengang and Chen, Dongdong and Chen, Dong and Yuan, Lu and Li, Houqiang},
  journal={IEEE Transactions on Pattern Analysis and Machine Intelligence},
  year={2025},
  publisher={IEEE}
}

@inproceedings{xia2024robust,
  title={Robust Cross-modal Medical Image Translation via Diffusion Model and Knowledge Distillation},
  author={Xia, Yuehan and Feng, Saifeng and Zhao, Jianhui and Yuan, Zhiyong},
  booktitle={2024 International Joint Conference on Neural Networks (IJCNN)},
  pages={1--8},
  year={2024},
  organization={IEEE}
}

@inproceedings{choi2020stargan,
  title={Stargan v2: Diverse image synthesis for multiple domains},
  author={Choi, Yunjey and Uh, Youngjung and Yoo, Jaejun and Ha, Jung-Woo},
  booktitle={Proceedings of the IEEE/CVF conference on computer vision and pattern recognition},
  pages={8188--8197},
  year={2020}
}

@article{karras2017progressive,
  title={Progressive growing of gans for improved quality, stability, and variation},
  author={Karras, Tero and Aila, Timo and Laine, Samuli and Lehtinen, Jaakko},
  journal={arXiv preprint arXiv:1710.10196},
  year={2017}
}

@article{choi2021ilvr,
  title={Ilvr: Conditioning method for denoising diffusion probabilistic models},
  author={Choi, Jooyoung and Kim, Sungwon and Jeong, Yonghyun and Gwon, Youngjune and Yoon, Sungroh},
  journal={arXiv preprint arXiv:2108.02938},
  year={2021}
}

@article{wang2004image,
  title={Image quality assessment: from error visibility to structural similarity},
  author={Wang, Zhou and Bovik, Alan C and Sheikh, Hamid R and Simoncelli, Eero P},
  journal={IEEE transactions on image processing},
  volume={13},
  number={4},
  pages={600--612},
  year={2004},
  publisher={IEEE}
}

@inproceedings{hu2022qs,
  title={Qs-attn: Query-selected attention for contrastive learning in i2i translation},
  author={Hu, Xueqi and Zhou, Xinyue and Huang, Qiusheng and Shi, Zhengyi and Sun, Li and Li, Qingli},
  booktitle={Proceedings of the IEEE/CVF Conference on Computer Vision and Pattern Recognition},
  pages={18291--18300},
  year={2022}
}

@inproceedings{brooks2023instructpix2pix,
  title={Instructpix2pix: Learning to follow image editing instructions},
  author={Brooks, Tim and Holynski, Aleksander and Efros, Alexei A},
  booktitle={Proceedings of the IEEE/CVF conference on computer vision and pattern recognition},
  pages={18392--18402},
  year={2023}
}

@article{hertz2022prompt,
  title={Prompt-to-prompt image editing with cross attention control},
  author={Hertz, Amir and Mokady, Ron and Tenenbaum, Jay and Aberman, Kfir and Pritch, Yael and Cohen-Or, Daniel},
  journal={arXiv preprint arXiv:2208.01626},
  year={2022}
}

@article{binkowski2018demystifying,
  title={Demystifying mmd gans},
  author={Bi{\'n}kowski, Miko{\l}aj and Sutherland, Danica J and Arbel, Michael and Gretton, Arthur},
  journal={arXiv preprint arXiv:1801.01401},
  year={2018}
}

@inproceedings{wu2023latent,
  title={A latent space of stochastic diffusion models for zero-shot image editing and guidance},
  author={Wu, Chen Henry and De la Torre, Fernando},
  booktitle={Proceedings of the IEEE/CVF International Conference on Computer Vision},
  pages={7378--7387},
  year={2023}
}

@article{kim2019u,
  title={U-gat-it: Unsupervised generative attentional networks with adaptive layer-instance normalization for image-to-image translation},
  author={Kim, Junho and Kim, Minjae and Kang, Hyeonwoo and Lee, Kwanghee},
  journal={arXiv preprint arXiv:1907.10830},
  year={2019}
}

@inproceedings{chen2020reusing,
  title={Reusing discriminators for encoding: Towards unsupervised image-to-image translation},
  author={Chen, Runfa and Huang, Wenbing and Huang, Binghui and Sun, Fuchun and Fang, Bin},
  booktitle={Proceedings of the IEEE/CVF conference on computer vision and pattern recognition},
  pages={8168--8177},
  year={2020}
}

@inproceedings{tumanyan2022splicing,
  title={Splicing vit features for semantic appearance transfer},
  author={Tumanyan, Narek and Bar-Tal, Omer and Bagon, Shai and Dekel, Tali},
  booktitle={Proceedings of the IEEE/CVF Conference on Computer Vision and Pattern Recognition},
  pages={10748--10757},
  year={2022}
}

@inproceedings{zhang2023modeling,
  title={Modeling spoof noise by de-spoofing diffusion and its application in face anti-spoofing},
  author={Zhang, Bin and Zhu, Xiangyu and Zhang, Xiaoyu and Lei, Zhen},
  booktitle={2023 IEEE International Joint Conference on Biometrics (IJCB)},
  pages={1--10},
  year={2023},
  organization={IEEE}
}

@inproceedings{popov2023optimal,
  title={Optimal transport in diffusion modeling for conversion tasks in audio domain},
  author={Popov, Vadim and Amatov, Amantur and Kudinov, Mikhail and Gogoryan, Vladimir and Sadekova, Tasnima and Vovk, Ivan},
  booktitle={ICASSP 2023-2023 IEEE International Conference on Acoustics, Speech and Signal Processing (ICASSP)},
  pages={1--5},
  year={2023},
  organization={IEEE}
}

@inproceedings{zhang2024decdm,
  title={DECDM: Document enhancement using cycle-consistent diffusion models},
  author={Zhang, Jiaxin and Rimchala, Joy and Mouatadid, Lalla and Das, Kamalika and Kumar, Sricharan},
  booktitle={Proceedings of the IEEE/CVF Winter Conference on Applications of Computer Vision},
  pages={8036--8045},
  year={2024}
}

@inproceedings{hur2024expanding,
  title={Expanding expressiveness of diffusion models with limited data via self-distillation based fine-tuning},
  author={Hur, Jiwan and Choi, Jaehyun and Han, Gyojin and Lee, Dong-Jae and Kim, Junmo},
  booktitle={Proceedings of the IEEE/CVF Winter Conference on Applications of Computer Vision},
  pages={5028--5037},
  year={2024}
}

@inproceedings{bourou2024phendiff,
  title={PhenDiff: Revealing subtle phenotypes with diffusion models in real images},
  author={Bourou, Anis and Boyer, Thomas and Gheisari, Marzieh and Daupin, K{\'e}vin and Dubreuil, V{\'e}ronique and De Thonel, Aur{\'e}lie and Mezger, Val{\'e}rie and Genovesio, Auguste},
  booktitle={International Conference on Medical Image Computing and Computer-Assisted Intervention},
  pages={358--367},
  year={2024},
  organization={Springer}
}

@inproceedings{yin2024scalable,
  title={Scalable motion style transfer with constrained diffusion generation},
  author={Yin, Wenjie and Yu, Yi and Yin, Hang and Kragic, Danica and Bj{\"o}rkman, M{\aa}rten},
  booktitle={Proceedings of the AAAI Conference on Artificial Intelligence},
  volume={38},
  number={9},
  pages={10234--10242},
  year={2024}
}

@article{mancusi2024latent,
  title={Latent Diffusion Bridges for Unsupervised Musical Audio Timbre Transfer},
  author={Mancusi, Michele and Halychanskyi, Yurii and Cheuk, Kin Wai and Moliner, Eloi and Lai, Chieh-Hsin and Uhlich, Stefan and Koo, Junghyun and Mart{\'\i}nez-Ram{\'\i}rez, Marco A and Liao, Wei-Hsiang and Fabbro, Giorgio and others},
  journal={arXiv preprint arXiv:2409.06096},
  year={2024}
}

@article{saharia2022image,
  title={Image super-resolution via iterative refinement},
  author={Saharia, Chitwan and Ho, Jonathan and Chan, William and Salimans, Tim and Fleet, David J and Norouzi, Mohammad},
  journal={IEEE transactions on pattern analysis and machine intelligence},
  volume={45},
  number={4},
  pages={4713--4726},
  year={2022},
  publisher={IEEE}
}

@inproceedings{mantiuk2012comparison,
  title={Comparison of four subjective methods for image quality assessment},
  author={Mantiuk, Rafa{\l} K and Tomaszewska, Anna and Mantiuk, Rados{\l}aw},
  booktitle={Computer graphics forum},
  volume={31},
  number={8},
  pages={2478--2491},
  year={2012},
  organization={Wiley Online Library}
}

@inproceedings{zhang2016colorful,
  title={Colorful image colorization},
  author={Zhang, Richard and Isola, Phillip and Efros, Alexei A},
  booktitle={Computer Vision--ECCV 2016: 14th European Conference, Amsterdam, The Netherlands, October 11-14, 2016, Proceedings, Part III 14},
  pages={649--666},
  year={2016},
  organization={Springer}
}

\clearpage
\section{Appendix}
\subsection{A. Additional references}\label{sec:More_related_works}
The application of DMs in I2I translation has experienced significant advancements, primarily due to their ability to generate high-quality and diverse outputs. Ho et al. \cite{su2022dual} introduced DDPMs, which achieved training stability through adopting a simplified noise scheduling strategy, thereby establishing a robust foundation for future investigations in this field. The introduction of Stable Diffusion \cite{rombach2022high} facilitated text-to-image generation by employing latent space compression and cross-attention mechanisms, making it a cornerstone for I2I translation. Subsequently, Saharia et al. developed Palette \cite{saharia2022palette}, a conditional diffusion framework that outperformed GANs in tasks such as colorization and image inpainting by utilizing source image conditioned iterative denoising. A brief overview of additional relevant research in this area follows:

\textbf{Unpaired I2I translation.} Unpaired I2I translation aims to learn cross-domain mappings without requiring paired data. The seminal work \cite{sasaki2021unit} introduced DMs to unpaired settings via joint distribution modeling. Su et al. \cite{su2022dual} employed independently trained DMs with PF-ODEs for translation, while Zhao et al. \cite{zhao2022egsde} used energy functions to preserve domain-invariant features. Other studies \cite{li2023bbdm,kim2024conditional,kim2023unpaired} treated translation as a stochastic Brownian bridge. Prompt-free methods \cite{parmar2023zero} leveraged editing directions and cross-attention guidance, and Tu et al. \cite{tu2024unpaired} improved diversity via diffusion adversarial networks and singular value decomposition (SVD)-based feature fusion. Luo et al. \cite{luo2024target} further introduced a target-guided strategy for unpaired cross-modal medical image translation.

\textbf{I2I translation in different fields.} (1) Medical Imaging:
Several technologies have been developed to preserve anatomical structures, including conditional adversarial diffusion \cite{xia2024diffusion}, contour-based representations \cite{chencontourdiff}, and frequency band separation \cite{li2023zero}. Graf et al. \cite{graf2023denoising} employed DMs to translate Magnetic Resonance Imaging into Computed Tomography images for automated spine segmentation. Wang et al. \cite{wang2024mutual} facilitated cross-modal alignment through interactive information guidance. Xing et al. \cite{xing2024cross} introduced cross-conditional DMs that leveraged target image distributions to enhance both the quality and efficiency of medical image translation. Patil et al. \cite{patil2025mmit} proposed a unified framework for single-source to multi-target translation in multi-modal medical imaging. (2) Remote Sensing and Sensors:
Brownian bridge processes have been applied to Synthetic Aperture Radar (SAR)–to–optical image translation \cite{guo2024learning,shi2024brain} and laser–to–visible light conversion \cite{zhang2025dcltv}. Vinholi et al. \cite{vinholi2024multi} enhanced the spatial resolution of optical images using DMs while ensuring radiometric consistency. Bai et al. sequentially incorporated SAR image conditioning \cite{bai2023conditional} and color information \cite{bai2024sar} into DMs to improve translation fidelity. (3) Video Processing:
Ceylan et al. \cite{ceylan2023pix2video} preserved temporal coherence by editing anchor frames and integrating self-attention features with depth conditioning.

\textbf{Structure preservation and Geometric alignment.} Ensuring geometric consistency is crucial in image translation. RePaint \cite{lugmayr2022repaint} preserved structural integrity during inpainting by iteratively sampling masked regions using unconditional DDPM priors. Lin et al. \cite{lin2023regeneration} maintained original content shapes in zero-shot I2I by leveraging rich cross-attention maps. Li et al. \cite{li2023injecting} mitigated structural distortion by integrating domain-independent geometric features with target appearances. Zheng et al. \cite{zheng2023layoutdiffusion} enhanced multi-object generation control through layout fusion modules and object-aware attention mechanisms. Han et al. \cite{han2024image} combined DMs with GPT architectures, enabling interpretable visual programming control and offering a novel approach to guiding image translation while preserving geometric fidelity. Finally, the study in \cite{lin2024mirrordiffusion} further improved geometric consistency by aligning text prompts and latent codes to mirror source and target images.

\textbf{Multi-modal and cross-modal generation.} Liu et al. \cite{liu2024iidm} leveraged style reference images and segmentation masks for conditional generation. Additionally, Gao et al. \cite{gao2024frequency} introduced frequency band separation to regulate text-guided translation processes. Sun et al. \cite{sun2023sddm} decomposed score functions on manifolds to optimize entangled distributions. Kim et al. \cite{kim2024adaptive} focused on translating 3D medical images from single- to multi-modal outputs. CRoSS \cite{yu2023cross} employed DDIM inversion for reversible steganography. Finally, Wang et al. \cite{wang2025sindiffusion} modeled patch distributions from single natural images to facilitate diverse image generation.

Beyond these categories, recent explorations include CRoSS \cite{yu2023cross} for reversible steganography and Robust Cross-Moda \cite{xia2024robust}, which improves adversarial robustness through knowledge distillation.
\subsection{B. Assumptions}\label{app:Assumptions}
We adopt the theoretical assumptions from \cite{song2021maximum,kwon2022score}. Notably, we present only the assumptions relevant to the source domain DM$^{A}$. The corresponding assumptions for the target domain DM$^{B}$ can be obtained analogously by replacing $^{A}$ with $^{B}$.
\begin{assumption}
    Both the drift term $ \boldsymbol{f}^{A}(\boldsymbol{x}^{A},t) $ and score network $ \boldsymbol{S}_{\boldsymbol{\theta}}^{A}(\boldsymbol{x}^{A},t) $ are Lipschitz continuous with respect to variable $ \boldsymbol{x}^{A} $, namely, there exist two scale continuous functions $ L_{\boldsymbol{f}}^{A}\left(t\right),L_{\boldsymbol{S}_{\boldsymbol{\theta}}}^{A}\left(t\right):\left[0,T\right]\to\mathbb{R}_{> 0} $, holding $ \| \boldsymbol{f}^{A}(\boldsymbol{x}^{A},t)-\boldsymbol{f}^{A}(\boldsymbol{y}^{A},t) \|_{2}\leq L_{\boldsymbol{f}}^{A}\left(t\right) \| \boldsymbol{x}^{A}-\boldsymbol{y}^{A} \|_{2}$ and $ \| \boldsymbol{S}_{\boldsymbol{\theta}}^{A}(\boldsymbol{x}^{A},t)-\boldsymbol{S}_{\boldsymbol{\theta}}^{A}(\boldsymbol{y}^{A},t) \|_{2}\leq L_{\boldsymbol{S}_{\boldsymbol{\theta}}}^{A}(t) \| \boldsymbol{x}^{A}-\boldsymbol{y}^{A} \|_{2}$ for any $ \boldsymbol{x}^{A},\boldsymbol{y}^{A}\in\mathbb{R}^{n} $, respectively. In addition, there exists constant $ C_{g}>0 $ for all $ t\in\left[0,T\right] $ such that $\frac{1}{C_{g}^{A}}<g^{A}\left(t\right)< C_{g}$.
\end{assumption}
\begin{assumption}
    There exists constant $ C_{p}>0 $ that satisfies $\{ \| \boldsymbol{S}_{\boldsymbol{\theta}}^{A}(\boldsymbol{x}^{A},t) \|_{2}, \| \nabla\log p_{t}^{A}(\boldsymbol{x}^{A}) \|_{2}\}\leq C_{p}(1+ \| \boldsymbol{x}^{A}  \|_{2})$ for any $ \boldsymbol{x}^{A}\in\mathbb{R}^{n} $ and $ t\in\left[0,T\right] $.
\end{assumption}
\begin{assumption}
    The score network $\boldsymbol{S}_{\boldsymbol{\theta}}^{A}(\boldsymbol{x}^{A},t)$ and probability density $p_{t}^{A}(\boldsymbol{x}^{A})$ satisfy conditions $ \{\int_{0}^{T}\mathbb{E}_{p_{t}^{A}}[ \| \boldsymbol{S}_{\boldsymbol{\theta}}^{A}(\boldsymbol{x}^{A},t)  \|_{2}^{2}]dt,\int_{0}^{T}\mathbb{E}_{p_{t}^{A}}[ \| \nabla\log p_{t}^{A}(\boldsymbol{x}^{A})  \|_{2}^{2}]dt\}<+\infty $.
\end{assumption}
\begin{assumption}
    There exists constant $ C_{e}>0 $ for all $ t\in\left[0,T\right] $ such that $ \{p_{t}^{A}(\boldsymbol{x}^{A}),q_{t}^{A}(\boldsymbol{x}^{A})\}=\mathcal{O}(\exp(-\| \boldsymbol{x}^{A} \|_{2}^{C_{e}})) $.
\end{assumption}
\begin{assumption}\label{assumption_song}
    We assume that DM$^{A}$ is precisely trained, i.e, $ \boldsymbol{S}_{\boldsymbol{\theta}}^{A}(\boldsymbol{x}^{A},t)\equiv\nabla\log p_{t}^{A}(\boldsymbol{x}^{A}) $ for all $t\in[0,T]$ and $\boldsymbol{x}^{A}\in\mathbb{R}^{n}$.
\end{assumption}
\subsection{C. Geometric variational principle of OT map}\label{Geometric variational principle of OT map}
Supposing $\mathcal{Y}^{B}=\{\boldsymbol{y}_{i}^{B}\}_{i\in\mathcal{I}}\overset{i.i.d}{\sim}p_{T}^{B}$ is the set of discrete points, here we introduce geometric variational methods \cite{gu2013variational} for calculating the OT map from $p_{T}^{A}$ to $\frac{1}{|\mathcal{I}|}\sum_{i\in\mathcal{I}}\delta ( \boldsymbol{y}^{B}-\boldsymbol{y}_{i}^{B} ) $. The dual of Monge problem (8) in main text is
\begin{equation}\label{Energy}
    \begin{split}
        &\max_{\psi}E(\psi)=\int_{\mathbb{R}^{n}}\psi^{c}(\boldsymbol{x}^{B})p_{T}^{A}(\boldsymbol{x}^{B})d\boldsymbol{x}^{B}+\frac{1}{|\mathcal{I}|}\sum_{i\in\mathcal{I}}\psi_{i}\\
        &=\sum_{i\in\mathcal{I}}\int_{W_{i}(\psi)}( \langle \boldsymbol{y}_{i}^{B},\boldsymbol{x}^{A}  \rangle-\psi_{i})d\boldsymbol{x}^{A}+\frac{1}{|\mathcal{I}|}\sum_{i\in\mathcal{I}}\psi_{i},
    \end{split}
\end{equation}
where $\psi_{i}=\psi(\boldsymbol{y}_{i}^{B})$, $\psi^{c}(\boldsymbol{x}^{A})=\min\limits_{i}\frac{1}{2}\|\boldsymbol{x}^{A}-\boldsymbol{y}_{i}^{B}\|_{2}^{2}-\psi_{i}$ and $W_{i}(\psi)=\{\boldsymbol{x}^{A}\in\mathbb{R}^{n}| \langle \boldsymbol{y}_{i}^{B}-\boldsymbol{y}_{j}^{B},\boldsymbol{x}^{A}  \rangle\geq\psi_{i}-\psi_{j},\forall j\in\mathcal{I}\}$ implies nearest power Voronoi diagram. Each target point $\boldsymbol{y}_{i}^{B}$ corresponds to a support plane $\pi_{i}= \langle \boldsymbol{x}^{A},\boldsymbol{y}_{i}^{B}  \rangle+\psi_{i}-\frac{1}{2}\|\boldsymbol{y}_{i}^{B}\|_{2}^{2}$, therefore the Brenier potential energy can be constructed as $u_{\boldsymbol{h}}(\boldsymbol{x}^{A})=\max\limits_{i} \langle \boldsymbol{x}^{A},\boldsymbol{y}_{i}^{B} \rangle+h_{i}$, here $h_{i}=\psi_{i}-\frac{1}{2}\|\boldsymbol{y}_{i}^{B}\|_{2}^{2}$. We optimize the energy function $E(\psi)$ in \eqref{Energy} through the Monte Carlo algorithm \cite{an2020ae} to obtain Brenier potential $u_{\boldsymbol{h}}$. Referring to this geometric property, the image of $\nabla u_{\boldsymbol{h}}$ falls precisely within the data set $\mathcal{Y}$, namely $\nabla u_{\boldsymbol{h}}(\boldsymbol{x}^{A})\in\mathcal{Y}$ for any $\boldsymbol{x}^{A}\sim p_{T}^{A}$. We denote the OT map from $p_{T}^{A}$ to $p_{T}^{B}$ as $u_{T}^{A\to B}$, so there is $u_{\boldsymbol{h}}\to u_{T}^{A\to B}$ when $|\mathcal{I}|\to+\infty$.
\subsection{D. Proofs}
In this section, we give a detailed derivation of the main theories proposed in this paper.
\subsubsection{D.1 Proof of Theorem 1}\label{app:theorem:Wasserstein_distance_upper_bound_DM}
\begin{proof}
    It can be directly proved based on the contraction property of Wasserstein distance \cite{carrillo2006contractions}    \begin{equation}\label{eq:contraction_property}
		\mathcal{W}_{2}(p_{T}^{B},q_{T}^{B})\leq\exp\left(-\frac{1}{2}\int_{0}^{T}L_{f}^{B}\left(t\right)dt\right)\mathcal{W}_{2}(p_{0}^{B},q_{0}^{B})
	\end{equation}
	and Theorem 1 in \cite{kwon2022score}. The scale function $L_{f}^{B}(t)$ in \eqref{eq:contraction_property} satisfies $ \lim\limits_{T\to\infty}\int_{0}^{T}L_{f}^{B}\left(t\right)dt=\infty $.
\end{proof}
\subsubsection{D.2 Proof of Theorem 2}\label{app:corollary:Wasserstein_distance_upper_bound_OUR}
\begin{proof}
    On the basis of Theorem 1 in \cite{kwon2022score}, we have
    \begin{equation}\label{eq:Corollary_3_1}
        \begin{split}
            &\mathcal{W}_{2}(p_{0}^{B},q_{0}^{B})\\
            \leq&\int_{0}^{T}\phi^{B}(t)^{\frac{1}{2}}\mathbb{E}_{p_{t}^{B}}[\| \nabla\log p_{t}^{B}(\boldsymbol{x}_{t}^{B})-\boldsymbol{S}_{\boldsymbol{\theta}}^{B}(\boldsymbol{x}_{t}^{B},t) \|_{2}^{2}]^{\frac{1}{2}}dt\\
            +&I(T)\mathcal{W}_{2}(p_{T}^{B},q_{T}^{B})\\
        \overset{\star}{\leq}&\sqrt{2T}\mathcal{J}_{SM}^{B}(\boldsymbol{\theta}^{B},\phi^{B},T)^{\frac{1}{2}}+I(T)\mathcal{W}_{2}(p_{T}^{B},q_{T}^{B})\\
        =&\sqrt{2T}(\mathcal{J}_{SM}^{B})^{\frac{1}{2}}+I(T)\mathcal{W}_{2}(\nabla u_{T}^{A\to B}(p_{T}^{A}),\nabla u_{\boldsymbol{h}}(p_{T}^{A})), 
        \end{split}
    \end{equation}
    where $\!I^{B}\left(T\right)=\exp\left(\int_{0}^{T}(L_{f}^{B}\left(t\right)+\frac{g^{B}\left(t\right)^{2}}{2}L_{\boldsymbol{S}_{\boldsymbol{\theta}}}^{B}\left(t\right))dt\right)\!$, $\star$ is derived from the Cauchy-Schwartz inequality and $\!\phi^{B}=(g^{B})^{4}(I^{B})^{2}\!$. 
    Referring to Proposition 2.1 of \cite{merigot2020quantitative} and Theorem 22 in \cite{chazal2017inference}, we concern with the last term of the above inequality in the following form	\begin{equation}\label{eq:Theorem_4}
		\begin{split}
            &\mathcal{W}_{2}(\nabla u_{T}^{A\to B}(p_{T}^{A}),\nabla u_{\boldsymbol{h}}(p_{T}^{A}))\\
			\leq&\| \nabla u_{T}^{A\to B}-\nabla u_{\boldsymbol{h}} \|_{L_{2}(p_{T}^{A})}\\
            =& \left(\int_{\mathbb{R}^{n}}\| \nabla u_{T}^{A\to B}(\boldsymbol{x}^{A})-\nabla u_{\boldsymbol{h}}(\boldsymbol{x}^{A}) \|_{2}^{2}p_{T}(\boldsymbol{x}^{A})d\boldsymbol{x}^{A}\right)^{\frac{1}{2}}\\
			\leq&\left(\int_{\mathbb{R}^{n}}\| \nabla u_{T}^{A\to B}(\boldsymbol{x}^{A})-\nabla u_{\boldsymbol{h}}(\boldsymbol{x}^{A}) \|_{2}^{2}d\boldsymbol{x}^{A}\right)^{\frac{1}{2}}\\
			\leq&2K_{1} \| u_{T}^{A\to B}-u_{\boldsymbol{h}}  \|_{\infty}^{\frac{1}{2}}(\| \nabla u_{T}^{A\to B}  \|_{\infty}^{\frac{1}{2}}+\left \| \nabla u_{\boldsymbol{h}} \right \|_{\infty}^{\frac{1}{2}})\\
			\leq& 4K_{1}K_{2} \| u_{T}^{A\to B}-u_{\boldsymbol{h}} \|_{\infty}^{\frac{1}{2}},
		\end{split}
	\end{equation}
	where $ K_{1} $ and $ K_{2} $ are constants, respectively. Then we can prove the below formula
    \begin{equation}
        \begin{split}
            \mathcal{W}_{2}(p_{0}^{B},q_{0}^{B})\leq&\sqrt{2T}(\mathcal{J}_{SM}^{B})^{\frac{1}{2}}+KI(T) \| u_{T}^{A\to B}-u_{\boldsymbol{h}} \|_{\infty}^{\frac{1}{2}}
        \end{split}
    \end{equation}
    by taking $ K=4K_{1}K_{2} $ and applying \eqref{eq:Theorem_4} to \eqref{eq:Corollary_3_1}.
\end{proof}
\subsubsection{D.3 Proof of Theorem 3}\label{app:theorem:Sample cyclic consistency}
\begin{proof}[Proof]
    All diffusion process can be described by a deterministic ODE with matching marginal distributions, known as the PF-ODE \cite{song2020score}, which uniquely encodes the data. For the forward SDE (1) in main text, the corresponding PF-ODE is shown as (3) in main text. In the reverse process, we replace $\nabla\log p_{t}^{A/B}(\boldsymbol{x}^{A/B})$ with the trained, parameterized score network $\boldsymbol{S}_{\boldsymbol{\theta}}^{A/B}(\boldsymbol{x}^{A/B},t)$ to obtain an approximate PF-ODE
    \begin{equation}\label{eq:probability_flow_ODE}
	d\boldsymbol{x}^{A/B}=(\boldsymbol{f}^{A/B}(\boldsymbol{x}^{A/B},t)-\frac{g^{A/B}(t)^{2}}{2}\boldsymbol{S}_{\boldsymbol{\theta}}^{A/B}(\boldsymbol{x}^{A/B},t))dt. 
\end{equation}
When $\eta=0$, the mapping from $\boldsymbol{x}_{t_{0}}^{A/B}$ to $\boldsymbol{x}_{t_{1}}^{A/B}$ then can be formulated as
\begin{equation}
\begin{split}
    &Solver_{\eta}(\boldsymbol{x}_{t_{0}}^{A/B},\boldsymbol{S}_{\boldsymbol{\theta}}^{A/B},t_{0},t_{1})\\
    =&\boldsymbol{x}_{0}^{A/B}+\int_{t_{0}}^{t_{1}}(\boldsymbol{f}^{A/B}(\boldsymbol{x}_{t}^{A/B},t)\\
    -&\frac{g^{A/B}(t)^{2}}{2}\boldsymbol{S}_{\boldsymbol{\theta}}^{A/B}(\boldsymbol{x}_{t}^{A/B},t))dt
\end{split}
\end{equation}
Under the premise that Assumption 5 is satisfied and the discretization error is ignored, $Solver_{\eta}$ completely follows  PF-ODE \eqref{eq:probability_flow_ODE}. Further, when the calculation error of OT is not taken into account, then we have $M_{ot,T}^{A\to B}\circ M_{ot,T}^{B\to A}(\boldsymbol{x}^{A/B})=\boldsymbol{x}^{A/B}$ for any $\boldsymbol{x}^{A/B}\in\mathbb{R}^{n}$. Based on the above analysis, OT-ALD satisfies the sample-level cycle consistency.
\end{proof}
\subsubsection{D.4 Proof of Theorem 4}\label{app:theorem:Distribution cyclic consistency}
\begin{proof}[Proof]
    Even if noise is introduced, namely, $\eta>0$, the marginal distribution $p_{t}^{A/B}$ always satisfies FPE (2) in main text. Under the premise that Assumption 5 is satisfied, and the discretization error and calculation error of OT are ignored, we have $q_{T}^{B}=M_{ot,T}^{A\to B}(p_{T}^{A})=p_{T}^{B}$ and $q_{T}^{A}=M_{ot,T}^{B\to A}(p_{T}^{B})=p_{T}^{A}$, which ensures that the forward FPE and approximate reverse FPE are equivalent \cite{kwon2022score}. Therefore, $q_{0}^{B}=p_{0}^{B}$, and further we have $p_{0}^{ A}=p_{0}^{\prime A}$ is satisfied. Based on the above analysis, OT-ALD satisfies the distributional-level cycle consistency.
\end{proof}

\subsection{E. More experimental details and visual results}
\subsubsection{E.1 More experimental details}\label{app:More experimental details}
\textbf{Hardware and Setup.} The experiments were conducted on an NVIDIA RTX 4090 cluster, utilizing CUDA 12.1 acceleration and implemented using the PyTorch framework.

\textbf{Codebase.} Our model is implemented based on the official GitHub repository from OpenAI (\url{https://github.com/openai/guided-diffusion}). All baseline models are reproduced using their respective official implementations: EGSDE is implemented using \url{https://github.com/ML-GSAI/EGSDE} with $T=1000$; CUT and CycleGAN use \url{https://github.com/taesungp/contrastive-unpaired-translation}; ILVR is based on \url{https://github.com/jychoi118/ilvr_adm} with $T=1000$; SDEdit uses \url{https://github.com/ermongroup/SDEdit} with $T=500$ and $K=2$ repetitions; and DDIB is implemented with \url{https://github.com/suxuann/ddib}. All hyperparameters follow the settings reported in the original papers.

\textbf{Implementation Parameters.} The OT-ALD introduced in this study is compared strictly in the same DDIM \cite{song2020denoising} backbone, employing a linear noisy schedule defined by the parameters ($\beta_{1}=1\times10^{-4}$, $\beta_{T}=0.02$, $T=1000$) for the AFHQ, CelebA-HQ and summer2winter dataset. More detailed hyperparameters are shown in Tab. \ref{tab:Optimization_hyper-parameters}. To construct the OT-ALD with truncation parameter $T_{trun}$, the initial values for $\beta_{1},\beta_{2},\cdots,\beta_{T}$ must first be established, followed by the cessation of the diffusion process at the $T_{trun}$-th step. To solve the Brenier potential energy $u_{\boldsymbol{h}}$ as delineated in (9) (main text), we employ the geometric variational principle, as referenced in Section \ref{Geometric variational principle of OT map} and \cite{an2020ae}. The number of Monte Carlo samples is established at $N_{mc}=10\times |\mathcal{I}|$. The learning rates for the datasets AFHQ, CelebA-HQ and summer2winter are set to $lr=1\times 10^{-5}$, $lr=1\times 10^{-5}$, and $lr=2\times 10^{-4}$, respectively. To enhance convergence, we double the number of samples $N_{mc}$ and reduce the learning rate $lr$ by a factor of 0.8 if the energy function $E(\boldsymbol{h})$ in (10) (main text) does not show a decrease over 50 iterations. Furthermore, a termination threshold $\tau$ is set to $8\times 10^{-4}$. The optimization procedure for $\boldsymbol{h}$ will cease when $E(\boldsymbol{h})<\tau$ or when the total number of iterations surpasses 10000.
\begin{table*}[htbp]
\caption{Optimization hyperparameters of OT-ALD on AFHQ, CelebA-HQ and summer2wnter.}\label{tab:Optimization_hyper-parameters}
\scalebox{1.04}{
\begin{tabular}{l|ccc}
\hline
 Hyperparameters& AFHQ-Cat 
($512\times 512$) & AFHQ-Dog 
($512\times 512$) & AFHQ-WIld 
($512\times 512$) \\ \hline
Noisy schedule &$\beta_1=1\times 10^{-4}$, $\beta_T=0.02$  &$\beta_1=1\times 10^{-4}$, $\beta_T=0.02$ &$\beta_1=1\times 10^{-4}$, $\beta_T=0.02$  \\
$T$ & 1000 & 1000 & 1000  \\
Number of residual blocks & 3 & 3 & 3\\
Model channels & 128 & 128 & 128\\
Channel multiple & (0.5, 1, 1, 2, 2, 4, 4) & (0.5, 1, 1, 2, 2, 4, 4) & (0.5, 1, 1, 2, 2, 4, 4)\\
Initial learning rate & $1\times 10^{-4}$ & $1\times 10^{-4}$ & $1\times 10^{-4}$ \\
Dropout & 0.0 & 0.0 & 0.0 \\
Adam optimizer $\beta_{1}^{A}$                         & 0.5 & 0.5 & 0.5 \\
Adam optimizer $\beta_{2}^{A}$                         & 0.9 & 0.9 & 0.9 \\
EMA                                     & 0.9999 & 0.9999 & 0.9999 \\
Batch size                              & 35 & 35 & 35 \\
Number of training steps               & 3850000 & 3640000 &3980000  \\
Number of GPUs               & 1$\times$RTX 4090 GPU & 1$\times$RTX 4090 GPU & 1$\times$RTX 4090 GPU \\ \hline
\end{tabular}}
\scalebox{1.04}{
\begin{tabular}{l|cc}
\hline
 Hyperparameters& CelebA-HQ-Male 
($512\times 512$) & CelebA-HQ-Female 
($512\times 512$)\\ \hline
Noisy schedule &$\beta_1=1\times 10^{-4}$, $\beta_T=0.02$  &$\beta_1=1\times 10^{-4}$, $\beta_T=0.02$\\
$T$ & 1000 & 1000\\
Number of residual blocks & 3 & 3\\
Model channels & 128 & 128\\
Channel multiple & (0.5, 1, 1, 2, 2, 4, 4) & (0.5, 1, 1, 2, 2, 4, 4)\\
Initial learning rate & $1\times 10^{-4}$ & $1\times 10^{-4}$\\
Dropout & 0.0 & 0.0\\
Adam optimizer $\beta_{1}^{A}$ & 0.5 & 0.5 \\
Adam optimizer $\beta_{2}^{A}$  & 0.9 & 0.9\\
EMA & 0.9999 & 0.9999\\
Batch size & 35 & 35\\
Number of training steps & 3820000 & 3770000\\
Number of GPUs               & 1$\times$RTX 4090 GPU & 1$\times$RTX 4090 GPU \\ \hline
\end{tabular}}\\
\scalebox{1.04}{
\begin{tabular}{l|cc}
\hline
 Hyperparameters& summer2winter-Summer ($256\times 256$) & summer2winter-Summer 
($256\times 256$)\\ \hline
Noisy schedule &$\beta_1=1\times 10^{-4}$, $\beta_T=0.02$  &$\beta_1=1\times 10^{-4}$, $\beta_T=0.02$\\
$T$ & 1000 & 1000\\
Number of residual blocks & 3 & 3\\
Model channels & 128 & 128\\
Channel multiple & (1, 1, 2, 2, 4, 4) & (1, 1, 2, 2, 4, 4)\\
Initial learning rate & $1\times 10^{-4}$ & $1\times 10^{-4}$\\
Dropout & 0.0 & 0.0\\
Adam optimizer $\beta_{1}^{A}$ & 0.5 & 0.5 \\
Adam optimizer $\beta_{2}^{A}$  & 0.9 & 0.9\\
EMA & 0.9999 & 0.9999\\
Batch size & 35 & 35\\
Number of training steps & 4370000 & 4460000\\
Number of GPUs               & 1$\times$RTX 4090 GPU & 1$\times$RTX 4090 GPU \\ \hline
\end{tabular}}
\end{table*}
\subsubsection{E.2 More visual results}
\begin{figure*} [htbp]
    \centering
    \renewcommand{\figurename}{Fig}
    \includegraphics[width=1.7\columnwidth]{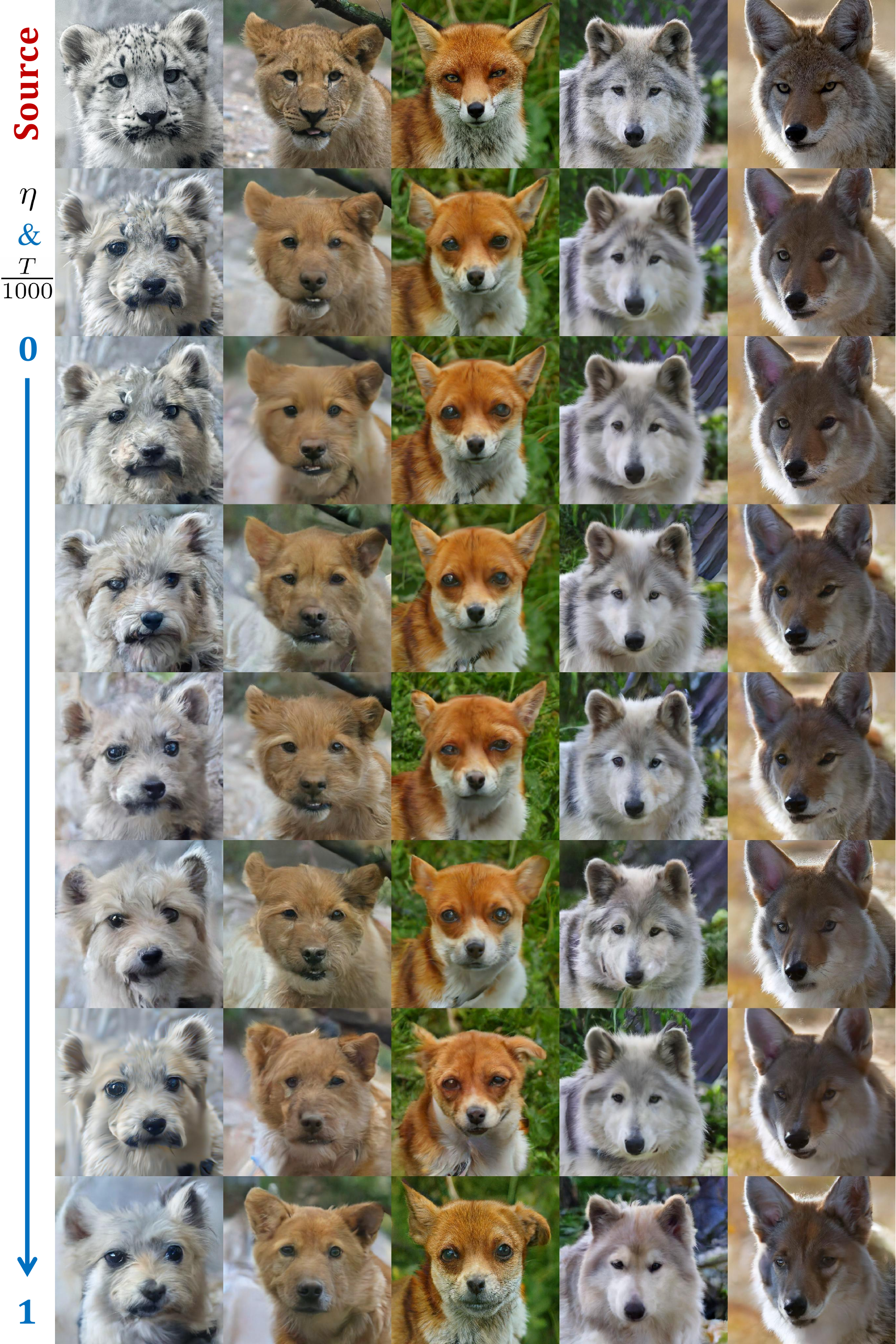}%
    \caption{Visualization of the translation results of OT-ALD with different $\eta$ and $T$ on Wild$\to$Dog task.}\label{framework2}
\end{figure*}

See Figs. \ref{framework}-\ref{framework2}.
\begin{figure*} [htbp]
    \centering
    \renewcommand{\figurename}{Fig}
    \includegraphics[width=1.7\columnwidth]{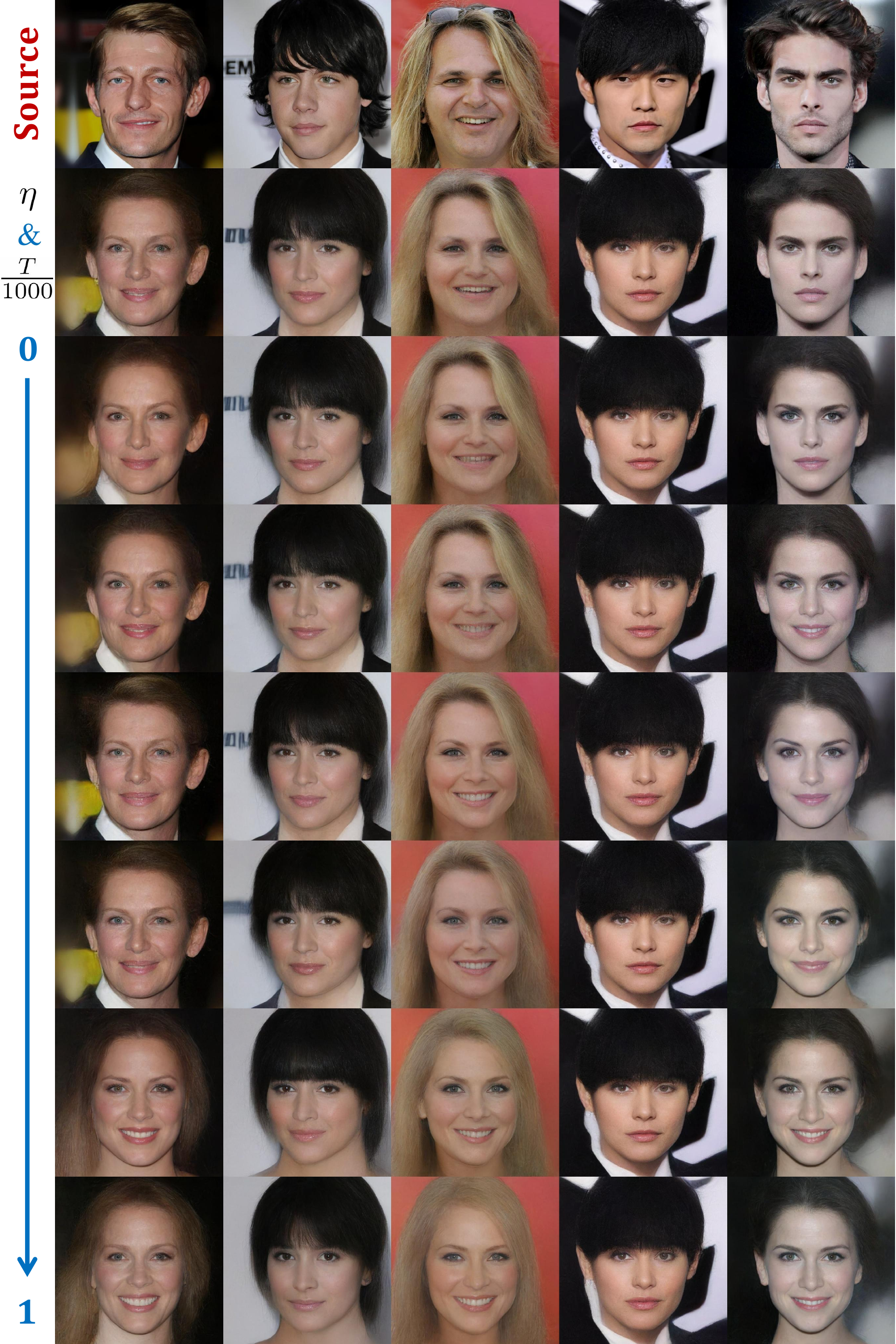}%
    \caption{Visualization of the translation results of OT-ALD with different $\eta$ and $T$ on Mela$\to$Female task.}\label{framework}
\end{figure*}
\begin{figure*} [htbp]
    \centering
    \renewcommand{\figurename}{Fig}
    \includegraphics[width=1.7\columnwidth]{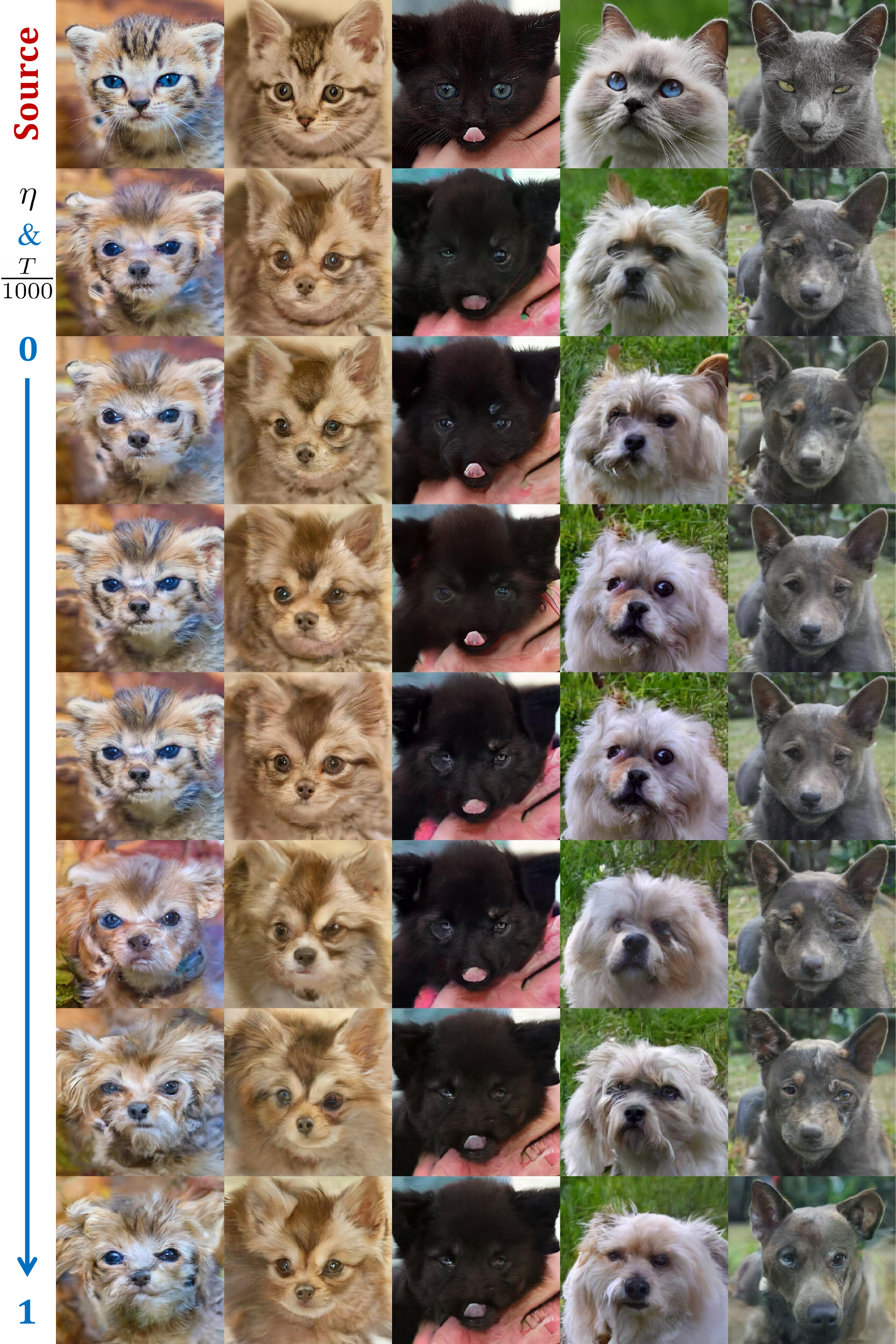}%
    \caption{Visualization of the translation results of OT-ALD with different $\eta$ and $T$ on Cat$\to$Dog task.}\label{framework1}
\end{figure*}
\end{document}